\documentclass[runningheads]{llncs}
\usepackage[T1]{fontenc}
\usepackage{graphicx}
\usepackage{booktabs}
\usepackage[misc]{ifsym}
\newcommand{\corr}{(\Letter)}
\usepackage{mwe}
\usepackage{graphicx}

\usepackage{tcolorbox}
\usepackage{xcolor}
\usepackage{multirow}
\usepackage{tabularx}
\usepackage{booktabs}
\usepackage{amsmath}
\usepackage{algorithm}      % For the algorithm environment
\usepackage{algpseudocode} % For the algorithmicx environment
\usepackage{adjustbox}
\usepackage{amssymb} % For \checkmark
\usepackage{tikz}
\usetikzlibrary{shapes.geometric, arrows}
\usepackage{fontawesome} % For icons
\usepackage{hyperref}
\usepackage{tcolorbox} % For the tcolorbox environment
\usepackage{fancyvrb} % For better verbatim control
\usepackage{caption} % For \ContinuedFloat
\usepackage{enumitem} % For better control over itemize

\begin{document}

\title{SCI-IDEA: Context-Aware Scientific Ideation\\ Using Token and Sentence Embeddings}

% \title{SCI-IDEA: Context-Aware Scientific Idea Generation using Token and Sentence Embeddings}

\titlerunning{SCI-IDEA}
% If the full title of your publication is short enough to also fit in the running head, you can omit the abbreviated publication title here. You can check as follows: if you comment out the \titlerunning line, something will appear in the header of all odd-numbered pages of your PDF from page 3 onward. This something is either the full title (in which case all is well), or the error message "Title Suppressed Due to Excessive Length". If this error message appears, you're going to want to provide an abbreviated title within the \titlerunning command, because if you won't do it, Springer will do it for you.

%N.B.: Author information (both in the \author{} and \authorrunning{} command) should only be present in the Camera-Ready Version of your publication. The version that you initially submit for review, ought to be double-blind. So, when initially submitting your publication, use:
%\author{Author information scrubbed for double-blind reviewing}

%........................................................................................
\author{Farhana Keya\inst{1} \and
Gollam Rabby\inst{2} \corr \and
Prasenjit Mitra\inst{1} \and
Sahar Vahdati\inst{1} \and \\
Sören Auer\inst{1,2} \and
Yaser Jaradeh\inst{2}} %\orcidID{0000-1111-2222-XXX}}
% You may leave out the orcidID information, if you want to.
% Use \corr to indicate the corresponding author. Note the spacing around the \corr command. Only one author can be the corresponding author.

%N.B.: comment out the \authorrunning{} command for the double-blind version of your publication submitted for review. Later, if your publication is accepted, use the command for the Camera-Ready Version.
\authorrunning{F. Keya et al.}
% First names are abbreviated in the running head.
% If there is one author, write 'A.L. Benjamin'.
% If there are two authors, write 'A.L. Benjamin and C.C. Broadus Jr.'
% If there are more than two authors, '[...] et al.' is used.

\institute{TIB—Leibniz Information Centre for Science and Technology, Hannover, Germany\email{\{farhana.keya, sahar.vahdati, auer\}@tib.eu}
\and
L3S Research Center, Leibniz University Hannover, Hannover, Germany \email{gollam.rabby, mitra, jaradeh @l3s.de}
% \and
% Secondary European Affiliation, Tiergartenstr. 17, 69121 Heidelberg, Germany
% \email{lncs@springer.com}  
\newline
 \vspace{0.2cm}
\raisebox{-1pt}{\faDatabase} \href{https://huggingface.co/datasets/tourist800/SCI-IDEA}{\texttt{Dataset}} \quad 
\raisebox{-1pt}{\faGlobe} \href{https://corei5.github.io/SCI-IDEA/}{\texttt{Project Page}} \quad 
\raisebox{-1pt}{\faGithub} \href{https://github.com/corei5/SCI-IDEA/tree/main}{\texttt{Codebase}}
}

%........................................................................................
\maketitle              % typeset the header of the contribution

\begin{abstract}

Every scientific discovery starts with an idea inspired by prior work, interdisciplinary concepts, and emerging challenges. Recent advancements in large language models (LLMs) trained on scientific corpora have driven interest in AI-supported idea generation. However, generating context-aware, high-quality, and innovative ideas remains challenging. 
We introduce SCI-IDEA, a framework that uses LLM prompting strategies and “Aha Moment” detection for iterative idea refinement. SCI-IDEA extracts essential facets from research publications, assessing generated ideas on novelty, excitement, feasibility, and effectiveness. 
Comprehensive experiments validate SCI-IDEA’s effectiveness, achieving average scores of 6.84, 6.86, 6.89, and 6.84 (on a 1–10 scale) across novelty, excitement, feasibility, and effectiveness, respectively. Evaluations employed GPT-4o, GPT-4.5, DeepSeek-32B (each under 2-shot prompting), and DeepSeek-70B (3-shot prompting), with token-level embeddings used for Aha Moment detection.
Similarly, it achieves scores of 6.87, 6.86, 6.83, and 6.87 using GPT-4o under 5-shot prompting, GPT-4.5 under 3-shot prompting, DeepSeek-32B under zero-shot chain-of-thought prompting, and DeepSeek-70B under 5-shot prompting with sentence-level embeddings. 
We also address ethical considerations such as intellectual credit, potential misuse, and balancing human creativity with AI-driven ideation. Our results highlight SCI-IDEA's potential to facilitate the structured and flexible exploration of context-aware scientific ideas, supporting innovation while maintaining ethical standards.

\end{abstract}


% %..................................................................
\vspace{-4mm}
\section{Introduction}
\label{sec:Introduction}

Ideation is the first step of any scientific project, involving the formulation of its purpose, research question, and hypotheses \cite{uhlmann2019scientific}. 
Scientists generate ideas at different rates due to epistemic motivation \cite{kruglanski2013lay}. Science of science has proved that social and collaborative environments also play a crucial role, as mentorship, thought collectives, and teamwork in boosting the chances of groundbreaking discoveries \cite{brorson2001stabilizing,brown2020language}. 
Not every scientist possesses all these traits at once. 
Large Language Models (LLMs) enhance scientific ideation by stimulating those factors mentioned above and beyond \cite{li2025review}. They act as virtual thought collectives, integrating multidisciplinary knowledge and historical breakthroughs. 
Additionally, LLMs and different agent-based frameworks serve as interactive brainstorming partners, refining hypotheses and fostering interdisciplinary collaboration, bridging gaps between research communities~\cite{gu2024interesting,papagni2023artificial}.
Recently, there has been growing attention on leveraging LLMs for scientific ideation \cite{si2024can}, yet existing methods often fail to balance novelty, relevance, and computational efficiency \cite{wang2023llmideation,baek2024researchagent,gu2024interesting}.
SCI-IDEA starts by extracting key facets from a researcher’s prior work or related literature to identify research gaps and opportunities for innovation.
The framework employs different prompting strategies based on researcher requirements: (1) Zero-shot prompting for straightforward context-aware idea generation, (2) Zero-shot chain-of-thought prompting for reasoning through multi-step research gaps, and (3) Few-shot prompting for tasks requiring domain-specific context. These strategies guide the LLM in generating context-aware scientific ideas while ensuring adaptability to diverse research queries.

Generated ideas are evaluated for novelty and surprise using semantic embedding methods to identify transformative ``Aha moments". \textbf{Novelty} is measured by semantic dissimilarity (cosine similarity on embeddings \cite{reimers2019sentence}).
\textbf{Surprise} is measured by the negative log-likelihood of ideas given their context, estimated using a pre-trained language model \cite{radford2019language}.
SCI-IDEA incorporates these metrics into a human-in-the-loop process, dynamically balancing idea exploration and refinement.

In such AI-assisted systems for science, evaluating the quality of generated text in general, and in this case, the quality of the generated ideas, remains a key challenge. 
In the SCI-IDEA framework, we have employed a systematic evaluation process for quality assessment of context-aware scientific ideas using four criteria: \textit{Novelty} (originality), \textit{Excitement} (engagement), \textit{Feasibility} (implementation potential), and \textit{Effectiveness} (goal achievement)~\cite{si2024can}. 
Human experts and LLMs independently score ideas on a 1-10 scale across these dimensions. This multi-dimensional evaluation aligns with recent AI-assisted ideation studies~\cite{si2024can}, ensuring the assessment of strengths and limitations. 

%\subsection{\textcolor{red}{Sahar: to be removed as subsection: Limitations of Existing Methods}}
\noindent \textbf{Related Work.} Based on the literature review that we conducted, existing methods for scientific ideation often rely on a single prompting strategy and fail to adapt to the iterative refinement of ideas ~\cite{wang2023llmideation,li2024learning,radensky2024scideator} and only focus on domain-driven tasks~\cite{gottweis2025towards,michinov2023creativity}. Additionally, traditional methods lack systematic mechanisms to evaluate the novelty and impact of generated ideas, often producing outputs that are either too conventional or irrelevant ~\cite{zhou2023creative,gu2024interesting,li2024learning}. 
SCI-IDEA addresses these limitations by dynamically evaluating novelty and surprise, enabling the identification of transformative ideas. It leverages insights from researchers' prior work and related literature to generate impactful ideas aligned with their expertise.

\begin{figure*}[htb]
    \centering
    \vspace{-4mm}
    \includegraphics[width=0.95\textwidth]{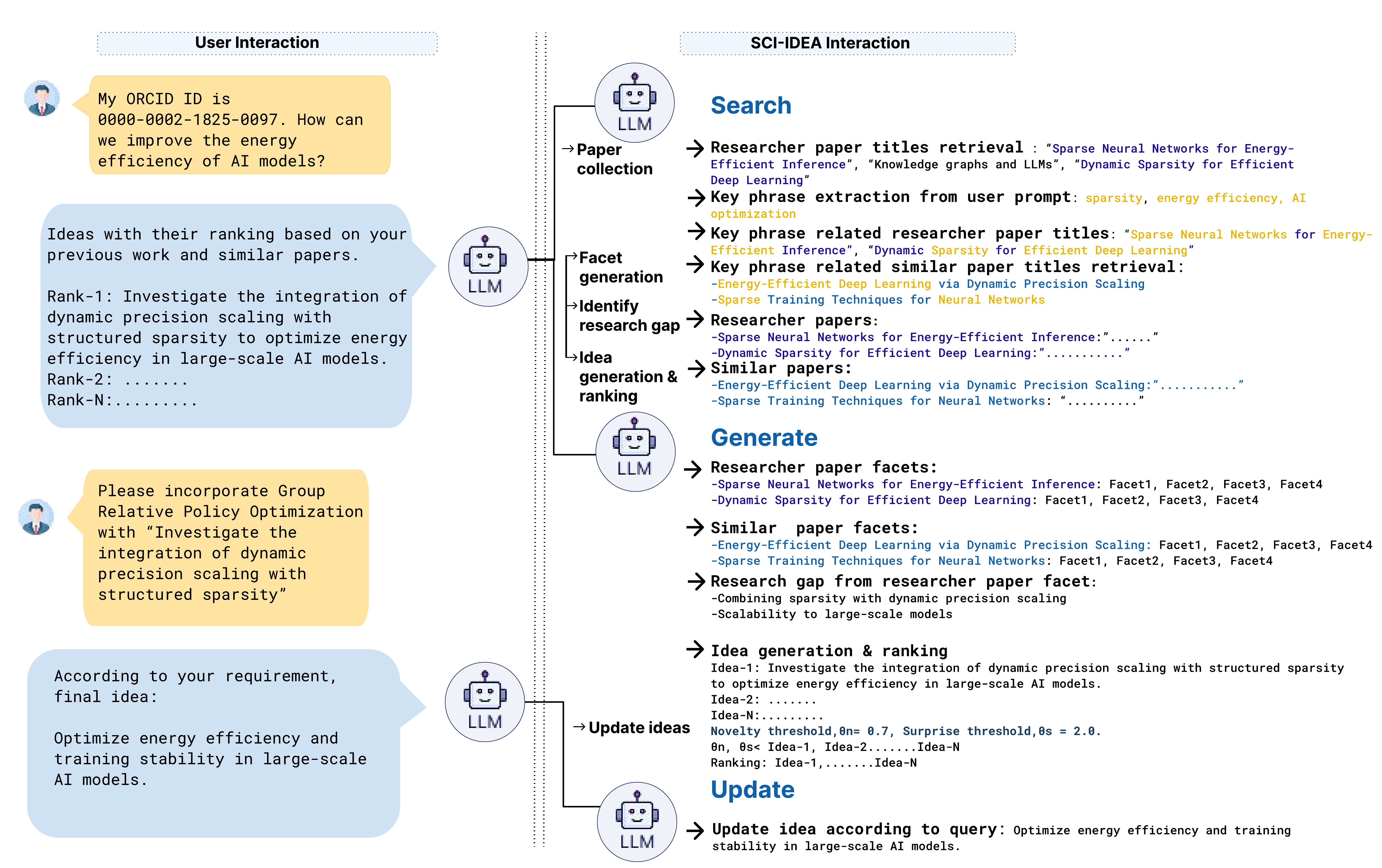} 
    \caption{\textbf{Overview of researcher and SCI-IDEA interactions}. The left side illustrates researcher interactions and feedback, while the right side highlights SCI-IDEA's techniques for generating and refining context-aware scientific ideas.}
    \label{fig:re_sci_conversation}
    \vspace{-5mm}
\end{figure*} 
\noindent \textbf{Motivating Example.} SCI-IDEA generates context-aware scientific ideas, such as: \textit{Using spiking neural networks (SNNs) with Group Relative Policy Optimization (GRPO) to optimize energy-efficient training of deep reinforcement learning (DRL) agents.} A researcher provides a query (e.g., \textit{How can we improve the energy efficiency of training DRL agents?}), and SCI-IDEA retrieves relevant publications from sources like arXiv and Semantic Scholar. Key facets—\textit{objectives (e.g., improving energy efficiency), methodologies (e.g., DRL, SNNs), and future work (e.g., hybrid DRL-SNN approaches)}—are extracted using LLMs. SCI-IDEA identifies research gaps, such as DRL's high energy consumption~\cite{mnih2015humanlevel} and SNNs' underutilization in reinforcement learning~\cite{lee2016training}~\cite{Nunes2022training}, forming the basis for idea generation. Using diverse prompting strategies, SCI-IDEA generates ideas. Zero-shot prompting explores energy-efficient alternatives to DRL, while zero-shot chain-of-thought prompting reasons step-by-step: (1) identifying energy-efficient architectures, (2) adapting them for reinforcement learning, and (3) applying them to agent control. Few-shot prompting provides examples (e.g., combining CNNs with reinforcement learning) to guide output. Ideas are evaluated for novelty (via cosine similarity~\cite{beltagy2019scibert}) and surprise (via negative log-likelihood ~\cite{radford2019language}). High novelty and surprise ideas are flagged as \textit{Aha moments}. Researcher feedback refines the idea, leading to: \textit{Using SNNs with GRPO to optimize energy-efficient training of DRL agents.} The final output is a ranked list of ideas, with the top idea summarized by its novelty (e.g., integrating SNNs, GRPO, and DRL), impact (e.g., reducing DRL energy consumption), and feasibility (e.g., supported by recent advancements)~\cite{lee2016training}~\cite{mnih2015humanlevel}. This demonstrates SCI-IDEA's ability to generate, evaluate, and refine high-impact scientific ideas through an iterative, human-in-the-loop process, as illustrated in \autoref{fig:re_sci_conversation}.

\vspace{-2mm}
\section{SCI-IDEA: Scientific Idea Generation Framework}
\label{sec:SCI-IDEA}

\vspace{-3mm}\subsubsection{Problem Formalization.}

Let us consider, given a researcher's scientific identifier (e.g., ORCID ID or institutional profile), the system accesses related research publications \( \mathcal{P} = \{p_1, p_2, \dots, p_n\} \) from academic sources such as CORE~\footnote{\url{https://core.ac.uk}}, arXiv~\footnote{\url{https://arxiv.org}}, and Semantic Scholar~\footnote{\url{https://semanticscholar.org}}. 
Each publication \( p_i \) is represented by its associated facets, including objectives, methodologies, evaluation, and future work. 
It is also possible that given a query \( q \), the system extracts keyphrases and retrieves a set of related publications \( \mathcal{A} = \{a_1, a_2, \dots, a_m\} \) with corresponding facets from the same sources. 
A researcher using the system can then select a subset of their publications \( \mathcal{P}_\text{selected} \subseteq \mathcal{P} \) to include in the ideation process. 
The goal is to generate a set of novel, impactful, and actionable scientific ideas \( \mathcal{I} = \{i_1, i_2, \dots, i_k\} \).

\vspace{-3mm}\subsubsection{Example of Researcher's Input and Facets.}
Let us assume the query of the researcher to SCI-IDEA is: \textit{How can we improve the energy efficiency of training deep reinforcement learning agents?} 
SCI-IDEA extracts keyphrases such as \emph{energy efficiency}, \emph{deep reinforcement learning}, and \emph{agents} to retrieve similar publications. 
Key facets extracted from the researcher's selected publications and similar literature include: 1) \textbf{Objectives}: Improving energy efficiency in agents. 2) \textbf{Methodologies}: Deep reinforcement learning (DRL), spiking neural networks (SNNs). 3) \textbf{Evaluation}: High energy consumption in DRL, potential of SNNs for low-power computation. 4) Future Work: Exploring hybrid approaches combining DRL and SNNs.

\vspace{-3mm}\subsubsection{Research Gap Identification.}
SCI-IDEA identifies research gaps by analyzing the structured facets extracted from a given profile and related research publications. 
For instance, it detects that while DRL is effective for agent control, its high energy consumption remains a significant limitation \cite{lee2016training}. Conversely, SNNs are known for their energy efficiency but have not been extensively applied to reinforcement learning. This gap serves as a foundation for generating research gaps. 
Our approach employs different prompting strategies in generating candidate ideas \( \mathcal{C} = \{c_1, c_2, \dots, c_p\} \) (Details can be found in the Appendix, Section: A). For example, using zero-shot chain-of-thought prompting, the underlying LLM might generate: \textit{Step 1: Identify energy-efficient neural architectures. Step 2: Adapt these architectures for reinforcement learning. Step 3: Apply them to agents' control tasks.} This process results in the research gap: \textit{How can spiking neural networks (SNNs) be effectively integrated into deep reinforcement learning (DRL) to optimize energy-efficient training while maintaining performance and stability?}

\vspace{-3mm}\subsubsection{Evaluation of Novelty and Surprise.}
The generated ideas are evaluated for two metrics of novelty and surprise, using state-of-the-art semantic embedding approaches. 
The novelty of an idea \( c_i \) is computed as:  
\[  
    \text{Novelty}(c_i) = 1 - \max(\text{Cosine\_similarity}(c_i, c_j)) \quad \forall c_j \in \mathcal{C}  
    \vspace{-1mm}
\]  
where \(\text{Cosine\_similarity}(c_i, c_j)\) is the cosine similarity between the semantic embeddings of \( c_i \) and \( c_j \). 
For example, if \( c_i \) is \textit{Using SNNs for energy-efficient DRL training} and \( c_j \) is \textit{Using CNNs for agents vision}, the low similarity score confirms its novelty among the generated ideas. Surprise of \( c_i \) is quantified as:  
\[  
    \text{Surprise}(c_i) = -\log p(c_i \mid \mathcal{C})  
\]  
where \( p(c_i \mid \mathcal{C}) \) is the likelihood of \( c_i \) given the context \( \mathcal{C} \), computed using a pre-trained language model~\cite{devlin2019bert,beltagy2019scibert,reimers2019sentence}. For example, the idea \textit{Using SNNs for energy-efficient DRL training} might have a high surprise score if it is unexpected yet reasonable. An idea is flagged as an Aha moment if it satisfies:  
\[  
\text{Novelty}(c_i) > \theta_n \quad \text{and} \quad \text{Surprise}(c_i) > \theta_s 
\]  
where \( \theta_n \) and \( \theta_s \) are predefined thresholds for novelty and surprise, respectively. Additionally, the idea must demonstrate novelty and surprise, such as proposing a novel, context-aware scientific idea.

\begin{figure*}[h!]
    \centering
    \includegraphics[width=0.95\textwidth]{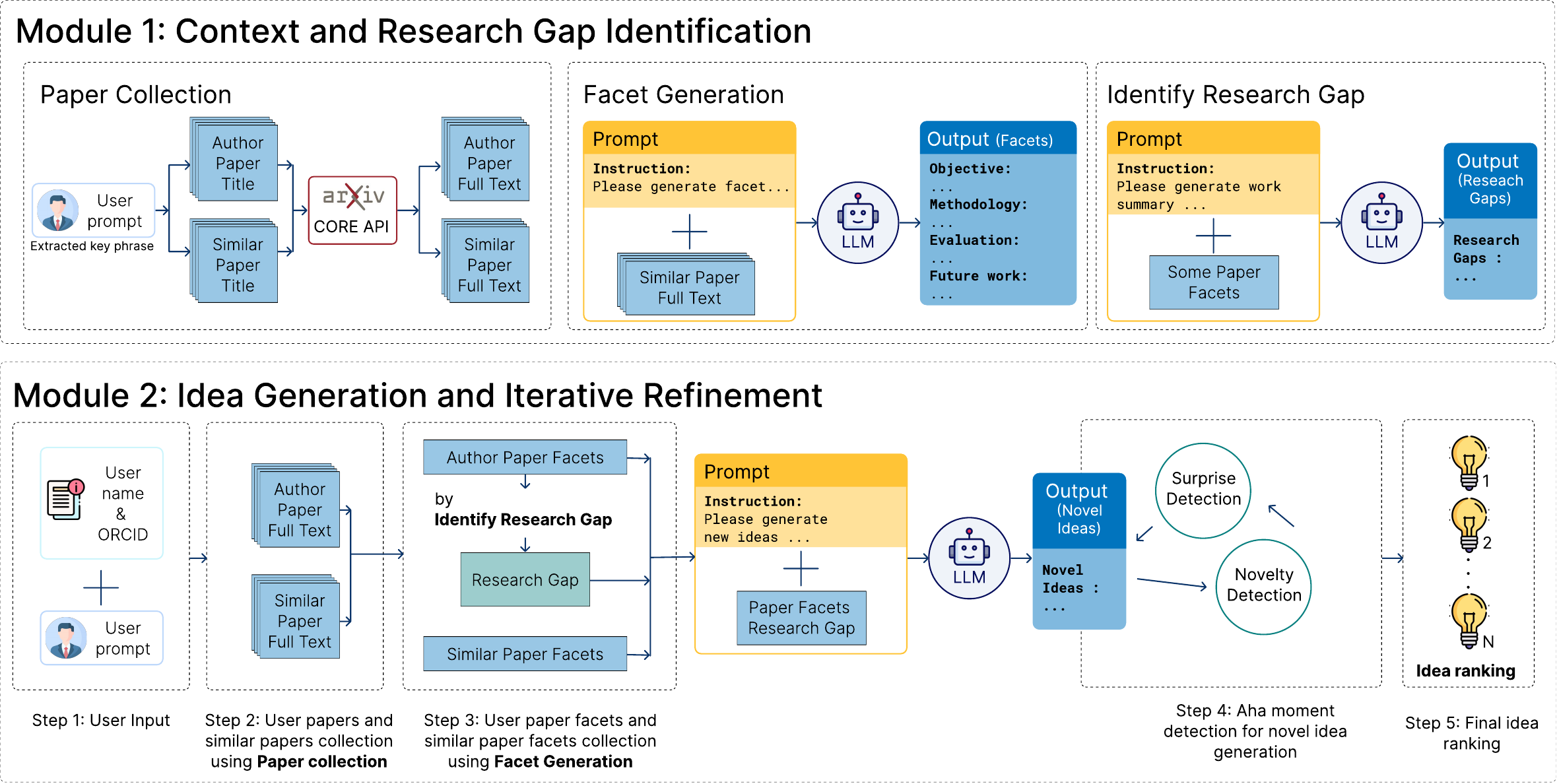} 
    \caption{\textbf{Overview of SCI-IDEA framework}. The upper side shows Module 1: Context Retrieval, Facet Extraction, and Research Gap Identification. The lower side illustrates Module 2: Idea Generation, Evaluation, Aha Moment Detection, and Refinement, along with their respective components.} 
    \label{fig:pipeline}
    \vspace{-3mm}
\end{figure*}

\vspace{-3mm}
\subsection{SCI-IDEA Framework Overview}

%..................................................................

SCI-IDEA is a framework for generating, evaluating, and refining context-aware scientific ideas through an iterative, human-in-the-loop process. 
The framework comprises two core modules: (1) \textit{Context and Research Gap Identification}, and (2) \textit{Idea Generation and Iterative Refinement}, as illustrated in \autoref{fig:pipeline}. The first module retrieves the publications of a given profile as well as related ones using a scientific identifier and research query. Key facets of these publications are extracted to identify research gaps, which serve as the foundation for generating context-aware scientific ideas using diverse prompting strategies (\autoref{sec:FERSRP}). (Details can be found in the Appendix, Section: A). The second module evaluates ideas for novelty and surprise using a pre-trained language model. Novelty is quantified by semantic dissimilarity, while the likelihood of the SCI-IDEA framework-generated ideas measures surprise. Ideas exceeding predefined thresholds for novelty and surprise are flagged as \textit{Aha moments}. The module enables iterative refinement, dynamically updating the context to balance exploration and exploitation, ensuring high-quality, context-aware results (\autoref{subsec:mod2}).

\subsection{Module 1: Context and Research Gap Identification}

\textbf{Context Retrieval.}
This process begins with a researcher providing their scientific identifier (e.g., ORCID ID: \texttt{0000-0002-1825-0097}) and a research query, such as: \textit{How can we improve the energy efficiency of AI models?}. Using the identifier, the system retrieves the researcher's publications, including their metadata and publicly available full-text. 
For example, the framework might retrieve the publication titled \textit{Sparse Neural Networks for Energy-Efficient Inference} from CORE for the above ID. 

\noindent \textbf{Context Extension.}
The framework uses an LLM to extract keyphrases representing the underlying research topic from the researcher query. For instance, the LLM extracts keyphrases such as \textit{sparsity}, \textit{energy efficiency}, and \textit{AI optimization} from the researcher's query. Building on the keyphrases extracted from the researcher's research query \( q \), the system retrieves related research publications from academic repositories. For example, for the researcher query the framework retrieves publications like \textit{Energy-Efficient Deep Learning via Dynamic Precision Scaling} from arXiv and \textit{Sparse Training Techniques for Neural Networks} from CORE, based on the keyphrases \textit{sparsity}, \textit{energy efficiency}, and \textit{AI optimization}.

\noindent \textbf{Facet Extraction.}
\label{sec:FERSRP}
Following the retrieval of full texts from publications authored by the researcher using the framework and related publications, the framework leverages an LLM to extract key facets, such as research objectives, methodologies, evaluation, and future work. For example, from the researcher's publication \textit{Sparse Neural Networks for Energy-Efficient Inference}, the framework extracts facets like the objective (\textit{reduce energy consumption in AI models}), methodology (\textit{structured sparsity techniques}), evaluation (\textit{30\% energy savings on benchmark datasets}), and future work (\textit{exploring hardware-software co-design for further energy optimization}). Similarly, from the related publication \textit{Energy-Efficient Deep Learning via Dynamic Precision Scaling}, the framework extracts facets such as the objective (\textit{optimize precision for energy efficiency}), methodology (\textit{dynamic precision scaling}), evaluation (\textit{20\% reduction in energy consumption with minimal accuracy loss}), and future work (\textit{extending the approach to real-time applications}). These facets provide a structured representation of the thematic and conceptual elements within the research content.

\noindent \textbf{Research Gap Identification.}
Using the structured facets extracted from the researchers and related research publications, the framework identifies research gaps by analyzing patterns, inconsistencies, and unexplored areas using the LLM. For instance, by comparing facets from \textit{Sparse Neural Networks for Energy-Efficient Inference} and \textit{Energy-Efficient Deep Learning via Dynamic Precision Scaling}, the framework highlights gaps such as \textit{underexplored methodologies} (e.g., combining sparsity with dynamic precision scaling) and \textit{unaddressed challenges} (e.g., scalability to large-scale models). The framework uses different prompting strategies to uncover these prospects depending on research requirements. For example, the framework might generate: \textit{Step 1: Identify energy-efficient sparsity and precision scaling techniques. Step 2: Explore potential synergies between these techniques. Step 3: Propose a hybrid approach combining sparsity and dynamic precision scaling.} This process ensures that the identified gaps are aligned with the researcher's interests.

\vspace{-3mm}
\subsection{Module 2: Idea Generation and Iterative Refinement}
\label{subsec:mod2}

% \vspace{-3mm}
\textbf{Idea Generation, Evaluation and Novelty with Surprise Detection.}

Building on identified research gaps, the framework generates context-aware scientific ideas using diverse prompting strategies. For example, based on gaps from the researcher's publication \textit{Sparse Neural Networks for Energy-Efficient Inference} and the similar publication \textit{Energy-Efficient Deep Learning via Dynamic Precision Scaling}, it proposes ideas such as: (1) \textit{integrating dynamic precision scaling with structured sparsity to optimize energy efficiency in large-scale AI models}, (2) \textit{using sparsity techniques to reduce energy consumption in AI models}, and (3) \textit{applying dynamic precision scaling for energy-efficient deep learning}. To evaluate novelty, the framework computes semantic embeddings of generated ideas and compares them using cosine similarity. For instance, the idea \textit{Investigate the integration of dynamic precision scaling and structured sparsity to optimize energy efficiency and training stability in large-scale AI models} is compared with embeddings of prior ideas, such as \textit{Use sparsity techniques to reduce energy consumption in AI models} and \textit{Apply dynamic precision scaling for energy-efficient deep learning}. A low maximum cosine similarity (e.g., 0.5) results in a high novelty score (e.g., 0.8), indicating a distinct and innovative idea. The novelty arises from uniquely combining sparsity and precision scaling. Surprise measures how unexpected an idea is given the context. For example, if the context focuses on sparsity techniques and the idea introduces dynamic precision scaling, the surprisal value will be high (e.g., 3.5), indicating an unexpected yet relevant insight. This occurs because the idea introduces a new dimension not emphasized by other ideas.

\begin{table*}[tb]
    \centering
    \scriptsize
    \caption{Template for zero-shot chain-of-thought (ZSCoT) with Aha moment flagging and deep dive exploration.}
    \label{tab:aha_template}
    \begin{tabular}{p{0.9\textwidth}}
        \hline
        
        \texttt{<\textcolor{blue!50!black}{think}> Let us think step by step. <\textcolor{blue!50!black}{/think}>} 
        \texttt{<\textcolor{blue!50!black}{answer}> Answer here....<\textcolor{blue!50!black}{/answer}>}....\texttt{<\textcolor{blue!50!black}{think}> \textcolor{blue!50!black}{Wait, this appears to be a breakthrough! We need a scientific idea .... multiple perspectives, and refinement for maximum impact.} <\textcolor{blue!50!black}{/think}>}  
        \texttt{<\textcolor{blue!50!black}{answer}> Refined answer here.........<\textcolor{blue!50!black}{/answer}>}  
        
        \texttt{<\textcolor{blue!50!black}{think}> \textcolor{blue!50!black}{An Aha moment has been detected! Let us delve deeper into this breakthrough idea.} <\textcolor{blue!50!black}{/think}>}  
        \texttt{<\textcolor{blue!50!black}{answer}>}  
        \textbf{\textcolor{blue}{\faRocket\ Deep Dive into a Breakthrough Idea}}  
        We have identified an \textbf{Aha moment!} Instead of stopping here, we will push further.  
        \textbf{How does this idea challenge conventional wisdom?}  
        \textbf....  
        \textbf{What are the top three practical applications of this idea?
        \textcolor{blue}{\faSearch\ Refining the Idea}}
        Let us refine this idea for \textbf{maximum novelty, feasibility, and impact}: .... \textcolor{blue}{\faLightbulbO\ Aha Moment Insight} 
        \texttt{<\textcolor{blue!50!black}{idea}>} {Idea here} \texttt{<\textcolor{blue!50!black}{/idea}>} \textcolor{blue}{\faFire\ Taking the Breakthrough Further}
        \texttt{<\textcolor{blue!50!black}{/answer}>}  
        \\
        \hline
    \end{tabular}
    \vspace{-5mm}
\end{table*}

\noindent \textbf{Difference Between Novelty and Surprise.}
An idea can be novel but not surprising if it is distinct from prior framework-generated ideas but aligns with expected research directions. For example, \textit{Use sparsity techniques to reduce energy consumption in AI models} may be novel if sparsity is unexplored in this context but unsurprising if sparsity is a well-known technique. An idea can be surprising but not novel if it is unexpected yet similar to existing ideas. For example, \textit{Apply dynamic precision scaling for energy-efficient deep learning} may be surprising in a sparsity-focused context but not novel if precision scaling is widely studied elsewhere. Also, an idea can be both novel and surprising if it introduces an unexpected and distinct combination of techniques. For example, \textit{Investigate the integration of dynamic precision scaling with structured sparsity to optimize energy efficiency in large-scale AI models} is both novel and surprising as it combines two distinct approaches. 

\noindent \textbf{Aha Moment Detection.}
An idea is flagged as an Aha moment if it satisfies the researcher's predefined thresholds for novelty and surprise (e.g., \( \theta_n = 0.7 \) and \( \theta_s = 2.0 \)). For example, the candidate idea above would qualify as an Aha moment if it introduced a novel combination of techniques (e.g., dynamic precision scaling with structured sparsity) and exceeded the thresholds for novelty and surprise.

\noindent \textbf{Iterative Exploration.}
When an idea is flagged as an Aha moment (e.g., \textit{Combining sparsity techniques with dynamic computation for energy-efficient AI models}), the Aha Prompt template (\autoref{tab:aha_template}) is activated to guide deeper exploration. The template directs the LLM to refine the idea from multiple perspectives, including feasibility, interdisciplinary insights, and practical applications. For instance, the system might refine the idea to: \textit{Explore hybrid architectures combining sparsity techniques with dynamic computation to reduce both parameter count and energy consumption during inference.} This iterative process continues until the generated ideas meet a researcher-selected threshold that the final outputs are not only novel and surprising but also actionable and aligned with the researcher's goals. By dynamically balancing the exploration of new conceptual spaces with the refinement of promising directions, the framework ensures the discovery of high-quality, context-aware scientific ideas.

\noindent \textbf{Final Idea Generation and Ranking.}
\label{subsubsec:final_ranking}
The generated candidate ideas are ranked using an LLM based on four criteria: \textit{Novelty}, \textit{Excitement}, \textit{Feasibility}, and \textit{Effectiveness}, ensuring a balanced evaluation of their impact and practicality. For example, for the query \textit{How can we improve the energy efficiency of training deep reinforcement learning agents?} and the identified gap between DRL's high energy consumption and SNNs' energy efficiency, the framework generates the idea: \textit{Develop a hybrid framework combining spiking neural networks (SNNs) with deep reinforcement learning (DRL) to optimize energy-efficient training while maintaining performance.} This idea scores high on novelty (bridging SNNs, DRL), excitement (addressing a critical challenge), and effectiveness (targeting the energy efficiency gap) but medium on feasibility due to the technical complexity of integration. In contrast, the idea \textit{Apply quantization techniques to reduce the precision of DRL parameters for energy savings} scores lower on novelty and excitement but higher on feasibility. The hybrid SNN and DRL idea is ranked higher due to its greater novelty, excitement, effectiveness, and Feasibility.

\noindent \textbf{Human-in-the-Loop Refinement.}
The last step of the framework includes human feedback where the researcher interacting with the system provides feedback, such as adding focus points. 
For instance, if the researcher suggests incorporating \textit{Group Relative Policy Optimization (GRPO)}, the framework updates the idea to: \textit{Investigate the integration of dynamic precision scaling, structured sparsity, and Group Relative Policy Optimization (GRPO) to optimize energy efficiency and training stability in large-scale AI models.} 
This human feedback process ensures the ideas are actionable, impactful, and more aligned with the researcher's interests. The system dynamically updates the context based on feedback, balancing exploration of ideas with exploitation of promising directions and fostering a collaborative and context-aware ideation process.

\begin{table}[tb]
\centering
\vspace{-4mm}
\caption{Keyphrase topic distribution - left and research profile wetrics (Extracted from Google Scholar) - right.}
\label{tab:idea_topic_distribution_and_research_metrics}
\scriptsize
\renewcommand{\arraystretch}{0.9} % Reduce row spacing
\setlength{\tabcolsep}{3pt} % Reduce column padding
\begin{minipage}{0.48\linewidth} % First table (left)
    \centering
    \begin{tabular}{lr}
    \toprule
    \textbf{Topic} & \textbf{Count} \\
    \midrule
    AI Alignment, RL, NLP & 36 \\
    Knowledge Representation & 12 \\
    Security, Privacy, Robustness in AI & 10 \\
    Materials Science, Chemistry & 17 \\
    Quantum Computing, Optimization & 8  \\
    Others & 17  \\
    \midrule
    \textbf{Total} & \textbf{100} \\
    \bottomrule
    \end{tabular}
\end{minipage}
\hfill % Creates spacing between the two tables
\begin{minipage}{0.48\linewidth} % Second table (right)
    \centering
    \begin{tabular}{l|r|r|r|r}
    \toprule
    \textbf{Stat} & \textbf{Articles} & \textbf{Cites} & \textbf{h-Index} & \textbf{i10} \\
    \midrule
    Mean   & 174  & 14,516  & 31  & 87  \\
    Median & 52   & 1,953   & 19  & 24.0  \\
    Min    & 2    & 1      & 1   & 0   \\
    Max    & 2,320 & 282,605 & 201 & 1,153 \\
    SD     & 343  & 37,219  & 38  & 188  \\
    \bottomrule
    \end{tabular}
\end{minipage}
\vspace{-3mm}
\end{table}
% \vspace{-10mm}

\section{Experimental Setup} 
\label{sec:experiments}

\textbf{Datasets and Baseline Models.}
\label{sec:datasets}
To evaluate SCI-IDEA, we curated a dataset of 100 researchers' profiles, including names, ORCID IDs, and publicly available publications. All researchers have published in top-tier computer science conferences (e.g., NeurIPS, ICML, ACL, SIGIR), ensuring high-impact relevance. For each researcher, we used a researcher prompt and an LLM to extract keyphrases, guiding the retrieval of similar research publications. The dataset covers diverse topics in computer science (\autoref{tab:idea_topic_distribution_and_research_metrics}), ensuring broad evaluation and generalizability. Three human annotators verified the dataset, cross-checking keyphrases and publication relevance. Research profile metrics (e.g., research publications, citations, h-index, i10-index) were computed for each researcher (\autoref{tab:idea_topic_distribution_and_research_metrics}), providing a comprehensive academic impact overview. While robust, the dataset has limitations, such as potential selection biases and reliance on Google Scholar data. To mitigate these, we ensured a balanced representation of senior and early-career researchers and cross-verified data from multiple sources, ensuring reliability and explanatory.

% \footnotesize
\begin{table}[ht]
    \centering
    \scriptsize % Further reduce table font size
    \caption{Performance of different LLMs without embeddings (without Aha moment).}
    \label{tab:without_embedding}
    \renewcommand{\arraystretch}{0.8} % Reduce row height
    \setlength{\tabcolsep}{2pt} % Reduce column spacing
    \resizebox{\columnwidth}{!}{ % Adjust table to fit within the page
    \begin{tabular}{p{1.5cm}|p{0.8cm}|p{1.2cm}|c|c|c|c|c}
        \hline
        \textbf{Model} & \textbf{Size} & \textbf{Prompt} & \textbf{Novelty} & \textbf{Excitement} & \textbf{Feasibility} & \textbf{Effectiveness} & \textbf{Avg} \\
        \hline
        GPT-4o & - & ZS & 7.10 & 7.67 & 5.84 & 6.91 & \textbf{6.88} \\
        GPT-4o & - & ZSCoT & 6.89 & 7.53 & 5.90 & 6.87 & 6.80 \\
        GPT-4o & - & 2FS & 6.80 & 7.37 & 5.70 & 6.73 & 6.65 \\
        GPT-4o & - & 3FS & 7.02 & 7.64 & 5.83 & 6.89 & 6.84 \\
        GPT-4o & - & 5FS & 6.95 & 7.59 & 5.84 & 6.92 & 6.82 \\
        \hline
        GPT-4.5 & - & ZS & 7.97 & 7.03 & 5.54 & 6.73 & 6.82 \\
        GPT-4.5 & - & ZSCoT & 7.91 & 6.98 & 5.42 & 6.56 & 6.72 \\
        GPT-4.5 & - & 2FS & 7.79 & 7.03 & 5.62 & 6.76 & 6.80 \\
        GPT-4.5 & - & 3FS & 7.90 & 7.03 & 5.75 & 6.84 & \textbf{6.88} \\
        GPT-4.5 & - & 5FS & 7.93 & 7.02 & 5.64 & 6.72 & 6.83 \\
        \hline
        DeepSeek & 32B & ZS & 6.93 & 6.88 & 6.07 & 6.63 & 6.63 \\
        DeepSeek & 32B & ZSCoT & 7.33 & 7.05 & 6.07 & 6.82 & 6.82 \\
        DeepSeek & 32B & 2FS & 7.29 & 7.31 & 6.00 & 6.88 & \textbf{6.87} \\
        DeepSeek & 32B & 3FS & 7.26 & 7.12 & 6.12 & 6.83 & 6.83 \\
        DeepSeek & 32B & 5FS & 7.34 & 7.07 & 6.00 & 6.76 & 6.79 \\
        \hline
        DeepSeek & 70B & ZS & 7.00 & 6.88 & 6.16 & 6.63 & 6.67 \\
        DeepSeek & 70B & ZSCoT & 7.09 & 7.15 & 6.14 & 6.83 & \textbf{6.80} \\
        DeepSeek & 70B & 2FS & 6.95 & 6.98 & 6.12 & 6.63 & 6.67 \\
        DeepSeek & 70B & 3FS & 7.18 & 7.00 & 6.13 & 6.79 & 6.78 \\
        DeepSeek & 70B & 5FS & 7.21 & 7.12 & 5.97 & 6.77 & 6.76 \\
        \hline
    \end{tabular}
    }
\end{table}

% \footnotesize % Reduce font size slightly
\begin{table}[ht]
    \centering
    \scriptsize % Further reduce table font size
    \caption{Performance of different LLMs with token-level embedding (with Aha moment).}
    \label{tab:token_level_embedding}
    \renewcommand{\arraystretch}{0.8} % Reduce row height
    \setlength{\tabcolsep}{2pt} % Reduce column spacing
    \resizebox{\columnwidth}{!}{ % Resize to fit within page
    \begin{tabular}{p{1.5cm}|p{1.2cm}|p{1.2cm}|c|c|c|c|c}
        \hline
        \textbf{Model} & \textbf{Size} & \textbf{Prompt} & \textbf{Novelty} & \textbf{Excitement} & \textbf{Feasibility} & \textbf{Effectiveness} & \textbf{Avg} \\
        \hline
        GPT-4o & - & ZS & 7.31 & 7.23 & 5.94 & 6.81 & 6.82 \\
        GPT-4o & - & ZSCoT & 7.19 & 7.12 & 6.14 & 6.85 & 6.82 \\
        GPT-4o & - & 2FS & 7.39 & 7.08 & 6.06 & 6.83 & \textbf{6.84} \\
        GPT-4o & - & 3FS & 7.38 & 7.09 & 5.89 & 6.75 & 6.78 \\
        GPT-4o & - & 5FS & 7.54 & 7.19 & 5.81 & 6.74 & 6.82 \\
        \hline
        GPT-4.5 & - & ZS & 7.71 & 7.00 & 5.84 & 6.84 & 6.85 \\
        GPT-4.5 & - & ZSCoT & 7.83 & 6.99 & 5.64 & 6.69 & 6.79 \\
        GPT-4.5 & - & 2FS & 7.68 & 7.02 & 5.93 & 6.81 & \textbf{6.86} \\
        GPT-4.5 & - & 3FS & 7.82 & 7.03 & 5.69 & 6.73 & 6.82 \\
        GPT-4.5 & - & 5FS & 7.84 & 7.01 & 5.71 & 6.73 & 6.82 \\
        \hline
        DeepSeek & 32B & ZS & 7.00 & 7.00 & 6.22 & 6.77 & 6.75 \\
        DeepSeek & 32B & ZSCoT & 7.32 & 7.15 & 6.04 & 6.89 & 6.85 \\
        DeepSeek & 32B & 2FS & 7.15 & 7.15 & 6.28 & 6.96 & \textbf{6.89} \\
        DeepSeek & 32B & 3FS & 6.88 & 7.01 & 6.17 & 6.72 & 6.69 \\
        DeepSeek & 32B & 5FS & 7.13 & 7.06 & 6.10 & 6.80 & 6.77 \\
        \hline
        DeepSeek & 70B & ZS & 6.81 & 7.12 & 6.23 & 6.84 & 6.75 \\
        DeepSeek & 70B & ZSCoT & 7.13 & 7.02 & 6.26 & 6.77 & 6.79 \\
        DeepSeek & 70B & 2FS & 6.07 & 7.12 & 6.71 & 7.13 & 6.76 \\
        DeepSeek & 70B & 3FS & 6.42 & 7.36 & 6.47 & 7.12 & \textbf{6.84} \\
        DeepSeek & 70B & 5FS & 6.76 & 7.24 & 6.34 & 6.96 & 6.82 \\
        \hline
    \end{tabular}
    }
    \vspace{-7mm}
\end{table}

For evaluating SCI-IDEA, we utilized GPT-4.5~\cite{openai2025gpt45}, GPT-4o~\cite{achiam2023gpt}, and DeepSeek-V3 (comprising DeepSeek-R1-Distill-Qwen-32B and DeepSeek-R1-Distill-Llama-70B)~\cite{guo2025deepseek}~\cite{liu2024deepseek}. These LLMs were chosen for their ability to generate high-quality scientific ideas and handle diverse prompts. GPT-4.5 and GPT-4o serve as general-purpose baselines, while DeepSeek-V3 leverages knowledge distillation for efficiency. We design prompts in three formats: zero-shot (ZS)~\cite{kojima2022large} for simplicity, few-shot (FS)~\cite{brown2020language} for context enrichment, and zero-shot chain-of-thought (ZSCoT)~\cite{kojima2022large} for step-by-step reasoning. We evaluate 2-shot, 3-shot, and 5-shot configurations using both closed-source (e.g., GPT-4.5, GPT-4o) and open-source (e.g., DeepSeek-V3) LLMs.

\noindent \textbf{Setup and Implementation.}
\label{subsec:evaluation_setups}

To assess the quality of generated scientific ideas, we employ a hybrid framework combining LLM-based and human expert evaluations. State-of-the-art LLMs rate each idea on four dimensions—novelty, excitement, feasibility, and expected effectiveness—using explicit evaluation instructions and reference examples to minimize biases and ensure reproducibility. Despite the efficiency of LLM-based evaluation, human judgment is essential for assessing deeper conceptual nuances. We recruit expert reviewers, primarily PhD holders with publications in top-tier conferences, to evaluate ideas using a structured review form mirroring the LLM criteria. Reviewers score ideas on a 1-10 scale and provide free-text rationales, with inter-rater reliability analyzed to ensure consistency. This hybrid approach ensures rigorous, high-quality evaluations that combine expert-level standards with the scalability of LLM-driven assessments.

% \footnotesize
\begin{table}[tb]
    \centering
    \vspace{-4mm}
    \scriptsize % Further reduce table font size
    \caption{Performance of different LLMs with sentence-level embedding (With Aha moment).}
    \label{tab:sentence_level_embedding}
    \renewcommand{\arraystretch}{0.8} % Reduce row height
    \setlength{\tabcolsep}{2pt} % Reduce column spacing
    \resizebox{\columnwidth}{!}{ % Resize to fit within page
    \begin{tabular}{p{1.5cm}|p{1.2cm}|p{1.2cm}|c|c|c|c|c}
        \hline
        \textbf{Model} & \textbf{Size} & \textbf{Prompt} & \textbf{Novelty} & \textbf{Excitement} & \textbf{Feasibility} & \textbf{Effectiveness} & \textbf{Avg} \\
        \hline
        GPT-4o & - & ZS & 7.76 & 6.99 & 5.67 & 6.71 & 6.78 \\
        GPT-4o & - & ZSCoT & 7.24 & 7.22 & 5.99 & 6.79 & 6.81 \\
        GPT-4o & - & 2FS & 7.30 & 7.06 & 6.07 & 6.78 & 6.80 \\
        GPT-4o & - & 3FS & 7.34 & 7.15 & 5.95 & 6.88 & 6.83 \\
        GPT-4o & - & 5FS & 7.61 & 7.13 & 5.94 & 6.81 & \textbf{6.87} \\
        \hline
        GPT-4.5 & - & ZS & 7.69 & 7.09 & 5.81 & 6.76 & 6.84 \\
        GPT-4.5 & - & ZSCoT & 7.78 & 6.98 & 5.55 & 6.65 & 6.74 \\
        GPT-4.5 & - & 2FS & 7.72 & 7.01 & 5.71 & 6.78 & 6.80 \\
        GPT-4.5 & - & 3FS & 7.71 & 7.00 & 5.84 & 6.87 & \textbf{6.86} \\
        GPT-4.5 & - & 5FS & 7.84 & 6.97 & 5.71 & 6.68 & 6.79 \\
        \hline
        DeepSeek & 32B & ZS & 7.08 & 6.88 & 6.15 & 6.67 & 6.69 \\
        DeepSeek & 32B & ZSCoT & 7.19 & 7.03 & 6.27 & 6.82 & \textbf{6.83} \\
        DeepSeek & 32B & 2FS & 7.08 & 6.91 & 6.12 & 6.82 & 6.73 \\
        DeepSeek & 32B & 3FS & 5.81 & 7.27 & 6.79 & 7.28 & 6.79 \\
        DeepSeek & 32B & 5FS & 6.92 & 7.09 & 6.27 & 6.90 & 6.79 \\
        \hline
        DeepSeek & 70B & ZS & 6.88 & 7.11 & 6.29 & 6.91 & 6.79 \\
        DeepSeek & 70B & ZSCoT & 6.97 & 6.95 & 6.31 & 6.79 & 6.76 \\
        DeepSeek & 70B & 2FS & 6.04 & 7.20 & 6.67 & 7.18 & 6.77 \\
        DeepSeek & 70B & 3FS & 5.27 & 7.25 & 7.06 & 7.52 & 6.78 \\
        DeepSeek & 70B & 5FS & 6.09 & 7.41 & 6.72 & 7.23 & \textbf{6.87} \\
        \hline
    \end{tabular}
    }
    \vspace{-7mm}
\end{table}

For semantic embeddings, we use two pre-trained transformer models: BERT (bert-base-uncased)~\cite{devlin2019bert} for general-purpose embeddings and SciBERT~\cite{beltagy2019scibert} for domain-specific scientific text. The text is tokenized and truncated to 512 tokens for efficiency, with embeddings computed as the mean across tokens. Novelty is calculated as \(1 - \max(\text{similarities})\) by comparing cosine similarities between new and previous responses, ensuring distinct and innovative ideas. All experiments were run on a GPU server, utilizing DeepSeek 70B for knowledge distillation and efficient idea generation and DeepSeek 32B for distillation, lightweight and scalable processing. The temperature settings for GPT-4o and DeepSeek were carefully chosen to balance creativity and consistency across tasks: for facet finding, GPT-4o and DeepSeek used temperature 0 and seed 1; for research gap identification, temperature 0.7 and seed 1; for idea generation, temperature 0.75 and seed 1; and for idea ranking, temperature 0.3 and seed 1. The temperature settings for GPT-4o were adopted from~\cite{peeperkorn2024temperature}, ensuring consistency with established practices in LLM-based ideation systems.

\vspace{-3mm}
\section{Results and Evaluation}
\label{sec:results}
\vspace{-2mm}

We analyze the SCI-IDEA performance across three variants: (1) without embeddings, (2) with token-level embeddings, and (3) with sentence-level embeddings. Each variant is evaluated under five prompting strategies: zero-shot, zero-shot chain-of-thought, and few-shot. Results are assessed across four metrics: novelty, excitement, feasibility, and effectiveness, with an overall average score computed for each configuration.

 \vspace{-2mm}
\subsection{Evaluation of Different Embeddings}

\vspace{-3mm}\subsubsection{Without Embeddings.}
The baseline without embeddings shows competitive performance in novelty and excitement (see~\autoref{tab:without_embedding}). 
GPT-4o achieves the highest average score of 6.88 using zero-shot prompting, with novelty and excitement scores of 7.10 and 7.67, respectively. However, this variant struggles with feasibility, as evidenced by lower scores across all LLMs (e.g., GPT-4o: 5.84, GPT-4.5: 5.75, DeepSeek-32B: 6.12). For example, GPT-4.5 achieves a feasibility score of only 5.55 under zero-shot chain-of-thought prompting, highlighting the challenge of generating practical ideas without embeddings. Despite these limitations, the baseline variant remains effective in generating novel and exciting ideas, with GPT-4.5 achieving an average score of 6.88 under 3-shot prompting.

\begin{figure*}[tb]
    % \vspace{-4mm}
    \centering
    \includegraphics[width=1.0\textwidth]{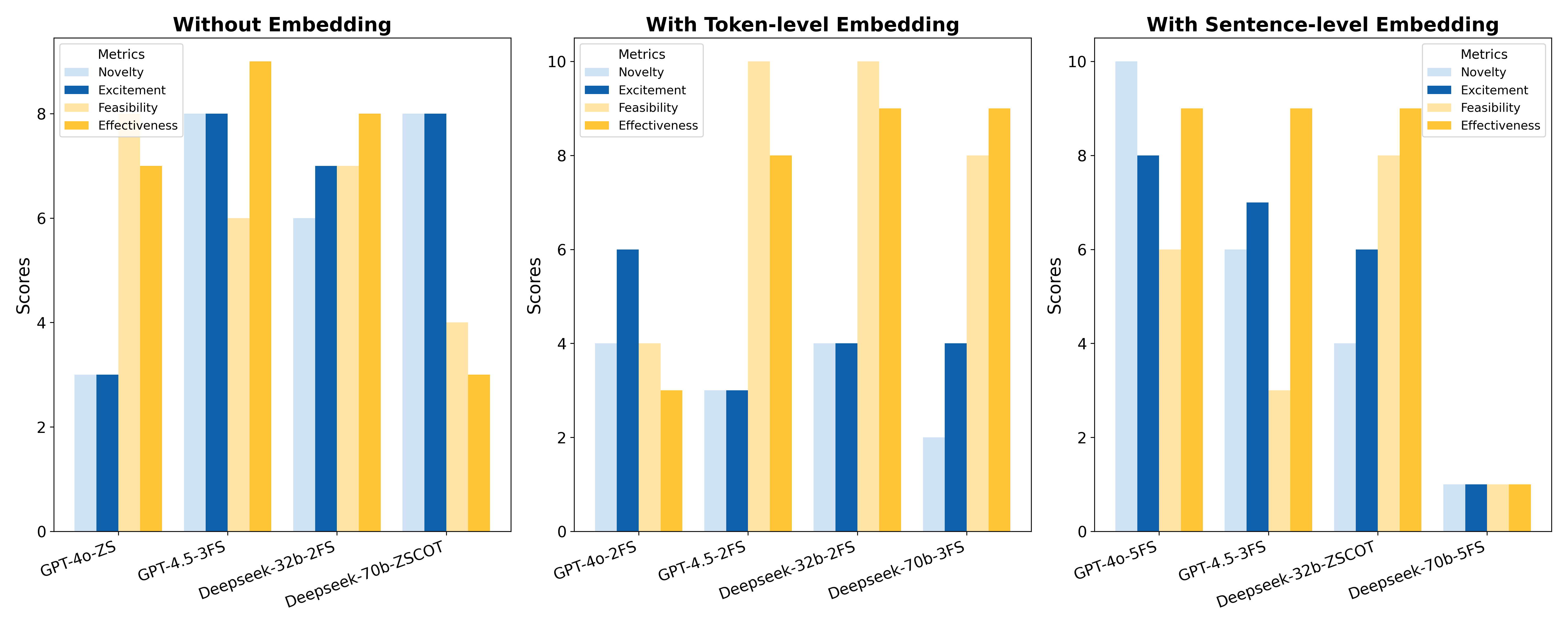} 
    \caption{\textbf{Human Evaluation of SCI-IDEA by Embedding Strategy}. Scores for novelty, excitement, feasibility, and effectiveness (left to right: without embedding, token-level embedding, sentence-level embedding).} 
    \label{fig:humaneva}
    % \vspace{-6mm}
\end{figure*} 

\begin{figure*}[tb]
    \centering
    \vspace{-8mm}
    \includegraphics[width=1.0\textwidth]{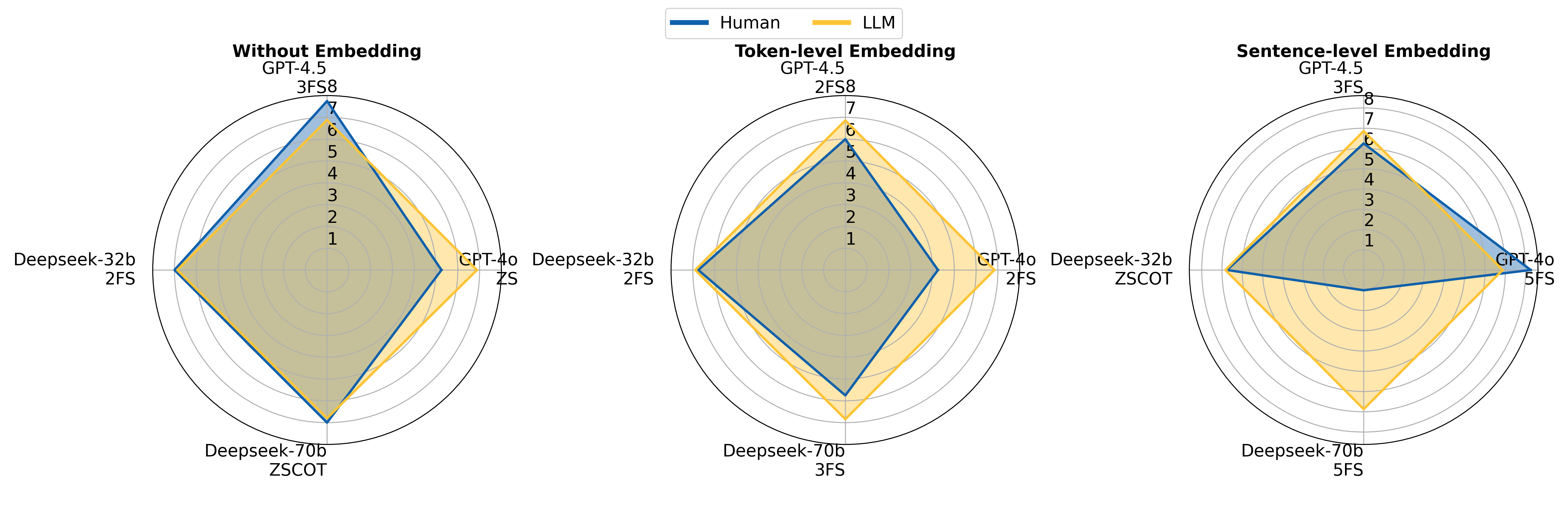} 
    \caption{\textbf{Comparison of Human vs. LLM Scores in SCI-IDEA}. Evaluation scores are across different prompting strategies (left to right: without embedding, token-level embedding, and sentence-level embedding).} 
    \label{fig:human_vs_LLM}
    \vspace{-6mm}
\end{figure*} 

\vspace{-3mm}\subsubsection{Token-Level Embeddings.}
The token-level embedding variant shows significant improvements in feasibility and effectiveness, as shown in~\autoref{tab:token_level_embedding}. DeepSeek-32B achieves the highest average score of 6.89 under 2-shot prompting, with feasibility and effectiveness scores of 6.28 and 6.96, respectively. This indicates that token-level embeddings enhance the practicality and relevance of generated ideas. GPT-4o and GPT-4.5 also perform well, with average scores of 6.84 and 6.86, respectively, under 2-shot prompting. GPT-4o achieves a feasibility score of 6.06 and an effectiveness score of 6.83 under 2-shot prompting, demonstrating the benefits of token-level embeddings in balancing novelty with practicality.

\vspace{-3mm}\subsubsection{Sentence-Level Embeddings.}
The sentence-level embedding variant demonstrates the best overall performance, particularly in novelty and excitement, as shown in~\autoref{tab:sentence_level_embedding}. GPT-4o achieves the highest average score of 6.87 under 5-shot prompting, with novelty and excitement scores of 7.61 and 7.13, respectively. This indicates that sentence-level embeddings are particularly effective in generating transformative scientific ideas. DeepSeek-70B also performs well, achieving an average score of 6.87 under 5-shot prompting, with novelty and excitement scores of 6.09 and 7.41, respectively. These results highlight the ability of sentence-level embeddings to produce unexpected and impactful ideas while maintaining high feasibility and effectiveness.

\vspace{-3mm}\subsubsection{Impact of Embeddings.}

The baseline without embeddings scores well in novelty and excitement but struggles with feasibility (e.g., GPT-4o: 5.84 feasibility, zero-shot prompting).
Token-level embeddings address this limitation, significantly improving feasibility and effectiveness. For instance, DeepSeek-32B achieves a feasibility score of 6.28 under 2-shot prompting with token-level embeddings, compared to 6.07 without embeddings. Sentence-level embeddings further enhance novelty and excitement, with GPT-4o achieving a novelty score of 7.61 under 5-shot prompting compared to 7.10 without embeddings. These results highlight the importance of embeddings in balancing novelty, feasibility, and effectiveness.

\vspace{-3mm}\subsubsection{Human Evaluation.}

We conducted a comprehensive human evaluation to assess SCI-IDEA's performance across prompting strategies and LLMs. Three evaluators with PhD scored outputs on four dimensions: \textit{Novelty}, \textit{Excitement}, \textit{Feasibility}, and \textit{Effectiveness} (\autoref{fig:humaneva}). For embeddings, GPT-4.5 with 3-shot prompting achieved the highest average score of 7.75, excelling in novelty and excitement, while DeepSeek-32B and DeepSeek-70B both scored 7 with 3-shot and zero-shot chain-of-thought prompting, respectively. In the token-level category, DeepSeek-32B with 2-shot prompting performed best with an average score of 6.75, demonstrating strong feasibility and effectiveness. For sentence-level embeddings, GPT-4o with 5-shot prompting outperformed others with an average score of 8.25, highlighting its ability to generate highly novel and effective ideas. Notably, DeepSeek-70B with 5-shot prompting underperformed significantly (score: 1 across all dimensions), indicating limitations in handling complex sentence-level tasks. These results demonstrate SCI-IDEA's effectiveness in leveraging diverse prompting strategies and LLMs, with GPT-4.5 and GPT-4o emerging as strong candidates for embedding and sentence-level tasks.

\vspace{-4mm}
\subsection{Ablation Studies}
\label{subsec:ablation}

\vspace{-3mm}\subsubsection{Bridging Human and LLM Evaluations.}
\autoref{fig:human_vs_LLM} compares human and LLM evaluation scores, revealing both alignment and divergence. For instance, GPT-4o with 5-shot prompting achieves the highest human score of 8.25 and a correspondingly high LLM score of 6.87, indicating strong agreement in evaluating novel and effective ideas. Similarly, DeepSeek-32b with 2-shot prompting shows consistent scores, with human evaluators assigning 6.75 and LLMs assigning 6.89, suggesting reliable performance. DeepSeek-70b with 5-shot prompting shows a significant difference, with human evaluators assigning a score of 1 compared to the LLM score of 6.87, highlighting LLMs' limitations in assessing low-quality or overly complex ideas. Conversely, GPT-4o with 2-shot prompting receives a moderate LLM score of 6.84 but a lower human score of 4.25, indicating LLMs may overestimate the quality of ideas lacking novelty or feasibility.

\vspace{-3mm}\subsubsection{Prompting Strategies.}
Few-shot prompting consistently outperforms zero-shot and zero-shot chain-of-thought prompting, particularly in feasibility and effectiveness. For example, DeepSeek-32B achieves its highest feasibility score of 6.28 with 2-shot prompting and token-level embeddings, compared to 6.07 under zero-shot prompting. Similarly, GPT-4.5 achieves its highest average score of 6.86 with 2-shot prompting and token-level embeddings, compared to 6.82 under zero-shot prompting. These results highlight the importance of providing context through examples, enhancing the LLM's ability to generate practical and relevant ideas.

\vspace{-3mm}\subsubsection{Performance with Smaller LLMs.}
To evaluate scalability, we test SCI-IDEA with smaller LLMs like DeepSeek-32B, comparing it against larger LLMs such as GPT-4o and GPT-4.5. Despite its reduced size, DeepSeek-32B achieves competitive performance, particularly with token-level embeddings, attaining an average score of 6.89 with 2-shot prompting. This is facilitated by knowledge distillation, which preserves reasoning capabilities while reducing computational overhead. However, smaller LLMs exhibit lower novelty scores, such as 7.00 under zero-shot prompting, compared to larger LLMs like GPT-4o, which achieves 7.61 under 5-shot prompting, reflecting a trade-off between LLM size and novelty generation. These results demonstrate SCI-IDEA's adaptability to resource-constrained environments while maintaining reasonable performance.

 \vspace{-2mm}
\section{Limitations and Ethical Considerations}

 \vspace{-2mm}
\subsubsection{Limitations of SCI-IDEA.}
\label{pra:ablation} 
While SCI-IDEA demonstrates significant advancements in context-aware scientific idea generation, it has several limitations. First, the framework relies heavily on the quality of input data, such as the researcher's and related publications, which may introduce biases or incomplete representations of the research landscape. Second, the evaluation metrics for novelty and surprise, while robust, are based on semantic embeddings and likelihood estimates, which may not fully capture the transformative potential of ideas. Third, the computational cost of iterative refinement and embedding-based evaluations can be prohibitive for large-scale applications. Additionally, the examples in this paper focus on computer science due to the authors' expertise, but the approach is broadly applicable to other domains.

\vspace{-3mm}\subsubsection{Ethical Considerations.}
The use of AI-generated scientific ideas raises ethical concerns, including misuse, intellectual credit ambiguity, and idea homogenization. Indiscriminate AI use could overwhelm academic venues and compromise peer-review integrity, necessitating that AI-assisted work meet rigorous human standards. LLM-assisted research also challenges traditional intellectual ownership, requiring standardized documentation of AI's role (e.g., LLMs, data sources, frameworks) for fair credit attribution. AI-generated ideas risk misuse, such as adversarial applications, highlighting the need for safety measures like Reinforcement Learning from Human Feedback (RLHF)~\cite{dai2023safe}. Additionally, current LLMs exhibit limited diversity in idea generation, raising concerns about the diversity of global research. Addressing this requires refining LLMs to encourage diverse perspectives and evaluate their ability to generate rare, transformative insights. Finally, AI integration into scientific ideation presents sociotechnical challenges, as over-reliance on AI could stifle human creativity. These issues must be managed to ensure responsible and ethical AI use in scientific ideation.

 \vspace{-3mm}
\section{Conclusion}
 \vspace{-3mm}

We introduced SCI-IDEA, a framework for generating context-aware, high-quality scientific ideas by leveraging the generative power of LLMs and structured evaluation metrics. SCI-IDEA addresses the limitations of existing methods by integrating iterative refinement, dynamic evaluation mechanisms, and domain-specific embeddings to ensure the novelty, excitement, feasibility, and effectiveness of generated ideas. Our experiments demonstrate that SCI-IDEA excels in context-aware ideation, achieving significant improvements in generating impactful and actionable research directions. We deem that SCI-IDEA has the potential to foster innovation and interdisciplinary collaboration across diverse scientific domains.
Future work will focus on addressing limitations by incorporating more diverse datasets, developing broader evaluation metrics, and optimizing computational efficiency. We also plan to explore the integration of domain-specific knowledge graphs and multi-modal inputs to enhance the quality and diversity of generated ideas. Additionally, an extensive human evaluation will be conducted to further validate the framework's effectiveness.

\vspace{-3mm}

\section*{Author Contributions}
Farhana Keya conducted the experiments and contributed to the manuscript writing. 
Gollam Rabby developed the initial idea, designed the experiments, and contributed to the manuscript writing. 
Prasenjit Mitra and Sahar Vahdati provided feedback on the initial idea and supported the manuscript writing. 
Sören Auer contributed to the initial idea and provided support in the manuscript writing. 
Yaser Jaradeh provided feedback on the manuscript and will integrate the idea into the Open Research Knowledge Graph (ORKG) alongside other related projects.

%Ameya and Andreas led the manuscript writing. Ludwig, Soren, Robert, Jenia and Matthias provided feedback. advice and scientific supervision throughout the project.

\section*{Acknowledgements}

The authors would like to thank Nahid Abdollahi for her insightful and valuable contributions to the visualization task. We also acknowledge the support of the KISSKI project (funding no. 01IS22093C) for providing computational resources, which will enable us to extend this research in the future.

% \section{Bib\TeX{} Files}
% \label{sec:bibtex}

\bibliographystyle{splncs04}  
\bibliography{Bibliography}

\newpage
\section*{A. Appendix}

% In the following sections, we report additional details on the following topics:

% \begin{itemize}
%     \item Examples of the SCI-IDEA system inputs, intermediate outputs, and final results (Section A.1)
%     \item All Unique Keys Found in SCI-IDEA Dataset (Section A.2)
%     \item Prompts in Experiment (Section A.3)
% \end{itemize}

In the following sections, we report additional details on the following topics:

\begin{enumerate}[leftmargin=*, noitemsep, topsep=0pt, partopsep=0pt]
    \item \textbf{All Unique Keys Found in SCI-IDEA Dataset} (Section A.1)
    \item \textbf{Prompts in Experiment} (Section A.2)
\end{enumerate}

% Add a horizontal rule for separation
\vspace{0.5cm} % Add some vertical space
\noindent\rule{\textwidth}{0.5pt} % Custom horizontal rule
\vspace{0.5cm} % Add some vertical space

\subsection*{A.1 All Unique Keys Found in SCI-IDEA Dataset}

\begin{tabular}{l|l}
\hline
\textbf{Column Name} & \textbf{Description} \\ \hline
Researcher Name & The name of the researcher. \\ \hline
ORCID & The unique ORCID identifier for the researcher. \\ \hline
Researcher Query Keyword & Keywords extracted from researcher synthetic query. \\ \hline
Research Full Paper & The full paper list for a researcher. \\ \hline
Similar Full Paper & List of full papers that are similar to the researcher query. \\ \hline
\end{tabular}

\subsection*{A.2 Prompts in Experiment}

\begin{figure}[htbp]
\centering
\begin{tcolorbox}[colback=blue!5!white, colframe=blue!75!black, title=Paper Facet Finder ZS, width=0.9\textwidth, boxrule=0.8mm, sharp corners=south]
\label{refinement-prompt}

\textbf{Paper:}

\vspace{0.5em} % Adjust this value to control spacing
\noindent
\textbf{Instruction:} Extract the following facets: Purpose, Mechanism, Evaluation, and Future Work.

\vspace{0.5em} % Adjust this value to control spacing
\noindent
\textbf{Format:}

\begin{Verbatim}[baselinestretch=0.8, fontsize=\small]
{
    Purpose: <subject> <predicate> <object>
    Mechanism: <subject> <predicate> <object>
    Evaluation: <subject> <predicate> <object>
    Future Work: <subject> <predicate> <object>
}
\end{Verbatim}
\end{tcolorbox}
\caption{Prompt for Paper Facet Finder ZS.} % Caption for the figure
\label{fig:prompt} % Label for cross-referencing
\end{figure}

\begin{figure}[htbp]
\centering
\begin{tcolorbox}[colback=blue!5!white, colframe=blue!75!black, 
    title=Paper facet finder 2FS, width=0.9\textwidth, 
    boxrule=0.8mm, sharp corners=south]

\textbf{Paper:} 

\vspace{0.5em} % Adjust this value to control spacing
\noindent
\textbf{Example 1} 

\vspace{0.5em} % Adjust this value to control spacing
\noindent
\textbf{Title:} A Novel Deep Learning Framework for Image Segmentation

\vspace{0.5em} % Adjust this value to control spacing
\noindent
\textbf{Facets:}
\begin{tcolorbox}[colback=white, colframe=black!50, width=\textwidth]
\small
\texttt{
\{
    Purpose: The paper proposes a deep learning framework for medical image segmentation. \\
    Mechanism: The framework utilizes a U-Net architecture with attention mechanisms. \\
    Evaluation: The model is evaluated using Dice score, IoU, and comparison with baseline models. \\
    Future Work: The authors suggest extending the model to 3D segmentation and multimodal learning.
\}
}
\end{tcolorbox}

\vspace{0.5em} % Adjust this value to control spacing
\noindent
\textbf{Example 2} 

\vspace{0.5em} % Adjust this value to control spacing
\noindent
\textbf{Title:} Knowledge Graph Embeddings for Recommendation Systems

\vspace{0.5em} % Adjust this value to control spacing
\noindent
\textbf{Facets:}
\begin{tcolorbox}[colback=white, colframe=black!50, width=\textwidth]
\small
\texttt{
\{
    Purpose: The paper explores knowledge graph embeddings for personalized recommendations. \\
    Mechanism: It leverages TransE and RotatE embeddings trained on interaction data. \\
    Evaluation: Performance is assessed using Hit@10, NDCG, and comparison with collaborative filtering. \\
    Future Work: The authors propose integrating user behavioral signals for better embeddings.
\}
}
\end{tcolorbox}

\vspace{0.5em} % Adjust this value to control spacing
\noindent
\textbf{Instruction:} Extract the following facets from the given paper: Purpose, Mechanism, Evaluation, and Future Work.

\vspace{0.5em} % Adjust this value to control spacing
\noindent
\textbf{Format:}
\begin{tcolorbox}[colback=white, colframe=black!50, width=\textwidth]
\small
\texttt{
\{
    Purpose: <subject> <predicate> <object> \\
    Mechanism: <subject> <predicate> <object> \\
    Evaluation: <subject> <predicate> <object> \\
    Future Work: <subject> <predicate> <object>
\}
}
\end{tcolorbox}

\end{tcolorbox}
\caption{Prompt for Paper Facet Finder 2FS.} % Caption for the figure
\label{fig:prompt} % Label for cross-referencing
\end{figure}

%..........................................................

\begin{figure}[htbp]
\centering
\begin{tcolorbox}[colback=blue!5!white, colframe=blue!75!black, 
    title=Paper facet finder 3FS, width=0.9\textwidth, 
    boxrule=0.8mm, sharp corners=south]

\textbf{Paper:} 

\vspace{0.5em} % Adjust this value to control spacing
\noindent
\textbf{Example 1} 

\vspace{0.5em} % Adjust this value to control spacing
\noindent
\textbf{Title:} Enhancing Speech Recognition with Deep Learning Integration

\vspace{0.5em} % Adjust this value to control spacing
\noindent
\textbf{Facets:}
\begin{tcolorbox}[colback=white, colframe=black!50, width=\textwidth ]
\small
\texttt{
\{
    Purpose: The study aims to improve speech recognition. \\
    Mechanism: The study integrates deep learning techniques into speech recognition. \\
    Evaluation: The study demonstrates improved accuracy in speech recognition. \\
    Future Work: The study suggests expanding the model to multilingual datasets.
\}
}
\end{tcolorbox}

\vspace{0.5em} % Adjust this value to control spacing
\noindent
\textbf{Example 2} 

\vspace{0.5em} % Adjust this value to control spacing
\noindent
\textbf{Title:} Improving Image Classification Accuracy with a Novel Algorithm

\vspace{0.5em} % Adjust this value to control spacing
\noindent
\textbf{Facets:}
\begin{tcolorbox}[colback=white, colframe=black!50, width=\textwidth ]
\small
\texttt{
\{
    Purpose: The paper presents a novel algorithm for image classification. \\
    Mechanism: The algorithm enhances accuracy by leveraging feature extraction. \\
    Evaluation: Experimental results show a 10\% increase in classification accuracy. \\
    Future Work: Future work includes adapting the algorithm for real-time applications.
\}
}
\end{tcolorbox}

\end{tcolorbox}
\caption{Prompt for Paper Facet Finder 3FS (Part 1).} % Caption for the first part
\label{fig:prompt} % Label for cross-referencing
\end{figure}

% Second part of the figure
\begin{figure}[htbp]
\centering
\begin{tcolorbox}[colback=blue!5!white, colframe=blue!75!black, 
    title=Paper facet finder 3FS (Continued), width=0.9\textwidth, 
    boxrule=0.8mm, sharp corners=south]

\vspace{0.5em} % Adjust this value to control spacing
\noindent
\textbf{Example 3} 

\vspace{0.5em} % Adjust this value to control spacing
\noindent
\textbf{Title:} Evaluating Model Performance on Benchmark Dataset: Significant Accuracy Improvements

\vspace{0.5em} % Adjust this value to control spacing
\noindent
\textbf{Facets:}
\begin{tcolorbox}[colback=white, colframe=black!50, width=\textwidth ]
\small
\texttt{
\{
    Purpose: The authors investigate the performance of the model. \\
    Mechanism: The model is applied to a benchmark dataset for validation. \\
    Evaluation: The model achieves a higher accuracy compared to previous approaches. \\
    Future Work: The authors plan to test the model on larger datasets.
\}
}
\end{tcolorbox}

\vspace{0.5em} % Adjust this value to control spacing
\noindent
\textbf{Instruction:} Extract the following facets from the given paper: Purpose, Mechanism, Evaluation, and Future Work.

\vspace{0.5em} % Adjust this value to control spacing
\noindent
\textbf{Format:}
\begin{tcolorbox}[colback=white, colframe=black!50, width=\textwidth ]
\small
\texttt{
\{
    Purpose: <subject> <predicate> <object> \\
    Mechanism: <subject> <predicate> <object> \\
    Evaluation: <subject> <predicate> <object> \\
    Future Work: <subject> <predicate> <object>
\}
}
\end{tcolorbox}

\end{tcolorbox}
\caption{Prompt for Paper Facet Finder 3FS (Part 2).} % Caption for the second part
\label{fig:prompt-cont} % Label for cross-referencing
\end{figure}

%..........................................................

\begin{figure}[htbp]
\centering
\begin{tcolorbox}[
    colback=blue!5!white, colframe=blue!75!black, 
    title=Paper facet finder 5FS, width=0.9\textwidth, 
    boxrule=0.8mm, sharp corners=south]

\textbf{Paper:} 

\vspace{0.5em} % Adjust this value to control spacing
\noindent
\textbf{Example 1} 

\vspace{0.5em} % Adjust this value to control spacing
\noindent
\textbf{Title:} Enhancing Speech Recognition with Deep Learning Integration

\vspace{0.5em} % Adjust this value to control spacing
\noindent
\textbf{Facets:}
\begin{tcolorbox}[colback=white, colframe=black!50, width=\textwidth ]
\small
\texttt{
\{
    Purpose: The study aims to improve speech recognition. \\
    Mechanism: The study integrates deep learning techniques into speech recognition. \\
    Evaluation: The study demonstrates improved accuracy in speech recognition. \\
    Future Work: The study suggests expanding the model to multilingual datasets.
\}
}
\end{tcolorbox}

\vspace{0.5em} % Adjust this value to control spacing
\noindent
\textbf{Example 2} 

\vspace{0.5em} % Adjust this value to control spacing
\noindent
\textbf{Title:} Improving Image Classification Accuracy with a Novel Algorithm

\vspace{0.5em} % Adjust this value to control spacing
\noindent
\textbf{Facets:}
\begin{tcolorbox}[colback=white, colframe=black!50, width=\textwidth ]
\small
\texttt{
\{
    Purpose: The paper presents a novel algorithm for image classification. \\
    Mechanism: The algorithm enhances accuracy by leveraging feature extraction. \\
    Evaluation: Experimental results show a 10\% increase in classification accuracy. \\
    Future Work: Future work includes adapting the algorithm for real-time applications.
\}
}
\end{tcolorbox}

\end{tcolorbox}
\caption{Prompt for Paper Facet Finder 5FS (Part 1).} % Caption for the first part
\label{fig:prompt} % Label for cross-referencing
\end{figure}

% Second part of the figure
\begin{figure}[htbp]
\ContinuedFloat % Continue the previous figure
\centering
\begin{tcolorbox}[
    colback=blue!5!white, colframe=blue!75!black, 
    title=Paper facet finder 5FS (Continued), width=0.9\textwidth, 
    boxrule=0.8mm, sharp corners=south]

\vspace{0.5em} % Adjust this value to control spacing
\noindent
\textbf{Example 3} 

\vspace{0.5em} % Adjust this value to control spacing
\noindent
\textbf{Title:} Evaluating Model Performance on Benchmark Dataset: Significant Accuracy Improvements

\vspace{0.5em} % Adjust this value to control spacing
\noindent
\textbf{Facets:}
\begin{tcolorbox}[colback=white, colframe=black!50, width=\textwidth ]
\small
\texttt{
\{
    Purpose: The authors investigate the performance of the model. \\
    Mechanism: The model is applied to a benchmark dataset for validation. \\
    Evaluation: The model achieves a higher accuracy compared to previous approaches. \\
    Future Work: The authors plan to test the model on larger datasets.
\}
}
\end{tcolorbox}

\vspace{0.5em} % Adjust this value to control spacing
\noindent
\textbf{Example 4} 

\vspace{0.5em} % Adjust this value to control spacing
\noindent
\textbf{Title:} Scaling the Model to Larger Datasets for Improved Generalization

\vspace{0.5em} % Adjust this value to control spacing
\noindent
\textbf{Facets:}
\begin{tcolorbox}[colback=white, colframe=black!50, width=\textwidth ]
\small
\texttt{
\{
    Purpose: The research explores model scalability. \\
    Mechanism: The model is designed to handle larger datasets. \\
    Evaluation: Preliminary tests indicate promising generalization performance. \\
    Future Work: The research suggests improving scalability for better generalization.
\}
}
\end{tcolorbox}

\end{tcolorbox}
\caption{Prompt for Paper Facet Finder 5FS (Part 2).} % Caption for the second part
\label{fig:prompt-cont} % Label for cross-referencing
\end{figure}

% Third part of the figure
\begin{figure}[htbp]
\ContinuedFloat % Continue the previous figure
\centering
\begin{tcolorbox}[
    colback=blue!5!white, colframe=blue!75!black, 
    title=Paper facet finder 5FS (Continued), width=0.9\textwidth, 
    boxrule=0.8mm, sharp corners=south]

\vspace{0.5em} % Adjust this value to control spacing
\noindent
\textbf{Example 5} 

\vspace{0.5em} % Adjust this value to control spacing
\noindent
\textbf{Title:} A Framework for Reducing Computational Cost While Preserving Accuracy

\vspace{0.5em} % Adjust this value to control spacing
\noindent
\textbf{Facets:}
\begin{tcolorbox}[colback=white, colframe=black!50, width=\textwidth ]
\small
\texttt{
\{
    Purpose: The framework optimizes computational efficiency. \\
    Mechanism: The framework reduces computation while preserving accuracy. \\
    Evaluation: The approach achieves a 30\% reduction in processing time. \\
    Future Work: Further enhancements will focus on energy-efficient implementations.
\}
}
\end{tcolorbox}

\vspace{0.5em} % Adjust this value to control spacing
\noindent
\textbf{Instruction:} Extract the following facets from the given paper: Purpose, Mechanism, Evaluation, and Future Work.

\vspace{0.5em} % Adjust this value to control spacing
\noindent
\textbf{Format:}
\begin{tcolorbox}[colback=white, colframe=black!50, width=\textwidth ]
\small
\texttt{
\{
    Purpose: <subject> <predicate> <object> \\
    Mechanism: <subject> <predicate> <object> \\
    Evaluation: <subject> <predicate> <object> \\
    Future Work: <subject> <predicate> <object>
\}
}
\end{tcolorbox}

\end{tcolorbox}
\caption{Prompt for Paper Facet Finder 5FS (Part 3).} % Caption for the third part
\label{fig:prompt-cont2} % Label for cross-referencing
\end{figure}

%......................................................................

\begin{figure}[htbp]
    \centering
    \begin{tcolorbox}[
        colback=blue!5!white, 
        colframe=blue!75!black, 
        title=Paper Facet Finder ZSCOT, 
        width=0.9\textwidth, 
        boxrule=0.8mm, 
        sharp corners=south
    ]

    \textbf{Paper:}

    \vspace{0.3em} % Reduced spacing
    \noindent
    \textbf{Instruction:} You are tasked with extracting four key facets from the given text: Purpose, Mechanism, Evaluation, and Future Work. \\
    Let's think step by step to ensure we identify each facet correctly.

    \vspace{0.3em} % Reduced spacing
    \noindent
    \textbf{Chain of Thought:}

    \begin{tcolorbox}[colback=white, colframe=black!50, width=\textwidth ]% Compact enumerate
    \small
        \textbf{Purpose}: This refers to the main objective or goal of the paper. To identify the purpose, ask yourself: What is the primary goal the author is trying to achieve? It often describes what the research aims to solve or improve.
        
        \textbf{Mechanism}: This is the method, process, or technique used to accomplish the purpose. Ask yourself: How is the goal being achieved? What is the approach, technique, or system being used?
        
        \textbf{Evaluation}: This describes how the effectiveness or success of the mechanism is assessed. Ask yourself: How was the approach tested or evaluated? What were the results?
        
        \textbf{Future Work}: This highlights the directions the author suggests for future research or improvements. Ask yourself: What does the author propose for future exploration or improvement in the field?
    \end{tcolorbox}

    \vspace{0.3em} % Reduced spacing
    \noindent
    \textbf{Format:}

    \begin{tcolorbox}[colback=white, colframe=gray, width=\textwidth ]
    \small
    \texttt{
    \{
        Purpose: <subject> <predicate> <object> \\
        Mechanism: <subject> <predicate> <object> \\
        Evaluation: <subject> <predicate> <object> \\
        Future Work: <subject> <predicate> <object>
    \}
    }
    \end{tcolorbox}

    \end{tcolorbox}
    \caption{Paper Facet Finder ZSCoT.}
\end{figure}

%..........................................................................

\begin{figure}[htbp]
\centering
\begin{tcolorbox}[
    colback=blue!5!white, colframe=blue!75!black, 
    title=Find research gap from author paper 2FS, width=0.9\textwidth, 
    boxrule=0.8mm, sharp corners=south ]

\label{work_summary-prompt}

\textbf{Paper:}
% \usepackage{tcolorbox}
% \tcbuselibrary{breakable}

\vspace{0.3em} % Reduced spacing
\noindent
\textbf{Paper <paper id>:}

\vspace{0.3em} % Reduced spacing
\noindent
\textbf{Title:} <paper title>

\vspace{0.3em} % Reduced spacing
\noindent
\textbf{Purpose:} <Purpose facet>

\vspace{0.3em} % Reduced spacing
\noindent
\textbf{Mechanism:} <Mechanism facet>

\vspace{0.3em} % Reduced spacing
\noindent
\textbf{Evaluation:} <Evaluation facet>

\vspace{0.3em} % Reduced spacing
\noindent
\textbf{Future Work:} <Future Work facet>

\vspace{0.5em} % Adjusted spacing
\noindent
\textbf{Instruction:}  
You are a Scientist, an intelligent assistant that helps researchers generate coherent, novel, and useful research ideas.  
Summarize the prior work above in approximately 200 words.  
Rather than summarizing individual papers one by one, summarize their contributions as a whole.  

Highlight:  
- What has already been done in research.  
- The research gaps from these prior works.  

This will help you avoid proposing redundant ideas.  
Instead, generate novel ideas that build upon the designated papers, leveraging their combined contributions.

\vspace{0.5em} % Adjusted spacing
\noindent
\textbf{Examples:}

\begin{itemize}[leftmargin=*, nosep, itemsep=0.3em] % Compact itemize
    \item \textbf{Example 1:}
    
    \textbf{Paper:}  
    \textbf{Title:} "Quantum Computing for Cryptography" \\
    \textbf{Purpose:} To enhance encryption security using quantum algorithms. \\
    \textbf{Mechanism:} Quantum key distribution (QKD) ensures secure communication. \\
    \textbf{Evaluation:} QKD resists decryption attempts by classical computers. \\
    \textbf{Future Work:} Research suggests improving scalability and error correction in quantum systems.

    \textbf{Answer:}  
    Quantum computing has revolutionized cryptography by leveraging quantum key distribution (QKD) to enhance security. Unlike classical methods, QKD provides theoretically unbreakable encryption. However, current implementations face scalability and error correction challenges. Future research should explore fault-tolerant quantum systems and practical large-scale deployment strategies to integrate QKD into real-world communication networks.
\end{itemize}

\end{tcolorbox}
\caption{Find research gap from author paper 2FS (Part 1).} % Caption for the first part
\label{fig:prompt} 
\end{figure}

% Second part of the figure
\begin{figure}[htbp]
\ContinuedFloat % Continue the previous figure
\centering
\begin{tcolorbox}[
    colback=blue!5!white, colframe=blue!75!black, 
    title=Find research gap from author paper 2FS (Continued), width=0.9\textwidth, 
    boxrule=0.8mm, sharp corners=south]

\begin{itemize}[leftmargin=*, nosep, itemsep=0.3em] % Compact itemize
    \item \textbf{Example 2:}
    
    \textbf{Paper:}  
    \textbf{Title:} "AI for Drug Discovery" \\
    \textbf{Purpose:} To accelerate the identification of potential drug candidates. \\
    \textbf{Mechanism:} Machine learning models predict molecular interactions for drug screening. \\
    \textbf{Evaluation:} AI-driven approaches reduce drug discovery time by 40\%. \\
    \textbf{Future Work:} Enhancing model interpretability and reducing false positives in predictions.

    \textbf{Answer:}  
    Artificial intelligence has significantly accelerated drug discovery by predicting molecular interactions with high efficiency, reducing the time required by 40\%. Despite these advances, AI models often lack interpretability and may produce false positives, leading to resource-intensive validation processes. Future work should focus on enhancing explainability and refining prediction accuracy through hybrid modeling approaches that integrate domain knowledge with machine learning techniques.
\end{itemize}

\vspace{0.5em} % Adjusted spacing
\noindent
\textbf{Now, summarize the contributions of the prior works above and identify research gaps:}

\vspace{0.5em} % Adjusted spacing
\noindent
\textbf{Format:}

\begin{tcolorbox}[colback=white, colframe=black!50, width=\textwidth ]
\small
\begin{Verbatim}[baselinestretch=0.8, fontsize=\small]
{
    Answer: 
}
\end{Verbatim}
\end{tcolorbox}

\end{tcolorbox}
\caption{Find research gap from author paper 2FS (Part 2).} % Caption for the second part
\label{fig:prompt-cont} % Label for cross-referencing
\end{figure}

%....................................................

\begin{figure}[htbp]
\centering
\begin{tcolorbox}[
    colback=blue!5!white, colframe=blue!75!black, 
    title=Find research gap from author paper 3FS, 
    width=0.9\textwidth, boxrule=0.8mm, sharp corners=south]

\label{work_summary-prompt}

\textbf{Paper:}

\vspace{0.3em} % Reduced spacing
\noindent
\textbf{Paper <paper id>:}

\vspace{0.3em} % Reduced spacing
\noindent
\textbf{Title:} <paper title>

\vspace{0.3em} % Reduced spacing
\noindent
\textbf{Purpose:} <Purpose facet>

\vspace{0.3em} % Reduced spacing
\noindent
\textbf{Mechanism:} <Mechanism facet>

\vspace{0.3em} % Reduced spacing
\noindent
\textbf{Evaluation:} <Evaluation facet>

\vspace{0.3em} % Reduced spacing
\noindent
\textbf{Future Work:} <Future Work facet>

\vspace{0.5em} % Adjusted spacing
\noindent
\textbf{Instruction:}  
You are a Scientist, an intelligent assistant that helps researchers generate coherent, novel, and useful research ideas.  
Summarize the prior work above in approximately 200 words.  
Rather than summarizing individual papers one by one, summarize their contributions as a whole.  

Highlight:  
- What has already been done in research?  
- The research gaps from these prior works.  
% \usepackage{tcolorbox}
% \tcbuselibrary{breakable}

This will help you avoid proposing redundant ideas.  
Instead, generate novel ideas that build upon the designated papers, leveraging their combined contributions.

\vspace{0.5em} % Adjusted spacing
\noindent
\textbf{Examples:}

\begin{itemize}[leftmargin=*, nosep, itemsep=0.3em] % Compact itemize
    \item \textbf{Example 1:}
    
    \textbf{Paper:}  
    \textbf{Title:} "Quantum Computing for Cryptography" \\
    \textbf{Purpose:} To enhance encryption security using quantum algorithms. \\
    \textbf{Mechanism:} Quantum key distribution (QKD) ensures secure communication. \\
    \textbf{Evaluation:} QKD resists decryption attempts by classical computers. \\
    \textbf{Future Work:} Research suggests improving scalability and error correction in quantum systems.

    \textbf{Answer:}  
    Quantum computing has revolutionized cryptography by leveraging quantum key distribution (QKD) to enhance security. Unlike classical methods, QKD provides theoretically unbreakable encryption. However, current implementations face scalability and error correction challenges. Future research should explore fault-tolerant quantum systems and practical large-scale deployment strategies to integrate QKD into real-world communication networks.
\end{itemize}

\end{tcolorbox}
\caption{Find research gap from author paper 3FS (Part 1).} % Caption for the first part
\label{fig:prompt} % Label for cross-referencing
\end{figure}

% Second part of the figure
\begin{figure}[htbp]
\ContinuedFloat % Continue the previous figure
\centering
\begin{tcolorbox}[
    colback=blue!5!white, colframe=blue!75!black, title=Find research gap from author paper 3FS (Continued), width=0.9\textwidth, boxrule=0.8mm, sharp corners=south]

\begin{itemize}[leftmargin=*, nosep, itemsep=0.3em] % Compact itemize
    \item \textbf{Example 2:}
    
    \textbf{Paper:}  
    \textbf{Title:} "AI for Drug Discovery" \\
    \textbf{Purpose:} To accelerate the identification of potential drug candidates. \\
    \textbf{Mechanism:} Machine learning models predict molecular interactions for drug screening. \\
    \textbf{Evaluation:} AI-driven approaches reduce drug discovery time by 40\%. \\
    \textbf{Future Work:} Enhancing model interpretability and reducing false positives in predictions.

    \textbf{Answer:}  
    Artificial intelligence has significantly accelerated drug discovery by predicting molecular interactions with high efficiency, reducing the time required by 40\%. Despite these advances, AI models often lack interpretability and may produce false positives, leading to resource-intensive validation processes. Future work should focus on enhancing explainability and refining prediction accuracy through hybrid modeling approaches that integrate domain knowledge with machine learning techniques.

    \item \textbf{Example 3:}
    
    \textbf{Paper:}  
    \textbf{Title:} "Edge Computing for Smart Cities" \\
    \textbf{Purpose:} To improve real-time data processing in urban environments. \\
    \textbf{Mechanism:} Edge devices process data locally, reducing cloud dependency. \\
    \textbf{Evaluation:} Edge computing lowers latency by 50\% compared to centralized systems. \\
    \textbf{Future Work:} Addressing security vulnerabilities and optimizing resource allocation in edge networks. %\usepackage{tcolorbox} \tcbuselibrary{breakable}

    \textbf{Answer:}  
    Edge computing has transformed smart city infrastructure by enabling real-time data processing, reducing latency by 50\% compared to centralized architectures. However, security vulnerabilities in edge networks remain a concern, as distributed systems are more susceptible to cyber threats. Additionally, optimizing resource allocation across edge devices presents a challenge. Future research should focus on developing robust security frameworks and dynamic resource management strategies to enhance edge computing reliability.
\end{itemize}

\vspace{0.5em} % Adjusted spacing
\noindent
\textbf{Now, summarize the contributions of the prior works above and identify research gaps:}
% \usepackage{tcolorbox}
% \tcbuselibrary{breakable}

\vspace{0.5em} % Adjusted spacing
\noindent
\textbf{Format:}
\begin{tcolorbox}[colback=white, colframe=black!50, width=\textwidth]
\small
\begin{Verbatim}[baselinestretch=0.8, fontsize=\small]
{
    Answer: 
}
\end{Verbatim}
\end{tcolorbox}

\end{tcolorbox}
\caption{Find research gap from author paper 3FS (Part 2).} % Caption for the second part
\label{fig:prompt-cont} % Label for cross-referencing
\end{figure}

%............................................................

\begin{figure}[htbp]
\centering
\begin{tcolorbox}[
    colback=blue!5!white, colframe=blue!75!black, 
    title=Find research gap from author paper 5FS, 
    width=0.9\textwidth, boxrule=0.8mm, sharp corners=south]

\label{work_summary-prompt}

\textbf{Paper:}

\vspace{0.3em} % Reduced spacing
\noindent
\textbf{Paper <paper id>:}

\vspace{0.3em} % Reduced spacing
\noindent
\textbf{Title:} <paper title>

\vspace{0.3em} % Reduced spacing
\noindent
\textbf{Purpose:} <Purpose facet>

\vspace{0.3em} % Reduced spacing
\noindent
\textbf{Mechanism:} <Mechanism facet>

\vspace{0.3em} % Reduced spacing
\noindent
\textbf{Evaluation:} <Evaluation facet>

\vspace{0.3em} % Reduced spacing
\noindent
\textbf{Future Work:} <Future Work facet>

\vspace{0.5em} % Adjusted spacing
\noindent
\textbf{Instruction:}  
You are a Scientist, an intelligent assistant that helps researchers generate coherent, novel, and useful research ideas.  
Summarize the prior work above in approximately 200 words.  
Rather than summarizing individual papers one by one, summarize their contributions as a whole.  

Highlight:  
- What has already been done in research?  
- The research gaps from these prior works.  

This will help you avoid proposing redundant ideas.  
Instead, generate novel ideas that build upon the designated papers, leveraging their combined contributions.

\vspace{0.5em} % Adjusted spacing
\noindent
\textbf{Examples:}

\begin{itemize}[leftmargin=*, nosep, itemsep=0.3em] % Compact itemize
    \item \textbf{Example 1:}
    
    \textbf{Paper:}  
    \textbf{Title:} "Deep Learning for Image Classification" \\
    \textbf{Purpose:} To enhance image classification accuracy. \\
    \textbf{Mechanism:} A deep convolutional neural network (CNN) architecture is used for feature extraction. \\
    \textbf{Evaluation:} The model achieves a 15\% improvement in classification accuracy over traditional methods. \\
    \textbf{Future Work:} Research suggests improving the network's generalization ability and adapting it to new datasets.

    \textbf{Answer:}  
    The research in deep learning for image classification focuses on improving accuracy through the use of convolutional neural networks (CNNs). Previous works have shown that CNNs significantly outperform traditional machine learning methods, with notable accuracy improvements of up to 15\%. However, these models often struggle with generalization to unseen datasets. Future work should focus on developing techniques to improve generalization, such as transfer learning or domain adaptation. This gap presents an opportunity for new methods to enhance the model's robustness across different types of data, allowing it to be more effective in real-world applications.

    \item \textbf{Example 2:} ...
\end{itemize}

\end{tcolorbox}
\caption{Find research gap from author paper 5FS (Part 1).} % Caption for the first part
\label{fig:prompt} % Label for cross-referencing
\end{figure}

% Second part of the figure
\begin{figure}[htbp]
\ContinuedFloat % Continue the previous figure
\centering
\begin{tcolorbox}[
    colback=blue!5!white, colframe=blue!75!black, 
    title=Find research gap from author paper 5FS (Continued), 
    width=0.9\textwidth, boxrule=0.8mm, sharp corners=south]

\begin{itemize}[leftmargin=*, nosep, itemsep=0.3em] % Compact itemize
    \item ...

    \item \textbf{Example 5:}
    
    \textbf{Paper:}  
    \textbf{Title:} "Blockchain for Secure Data Sharing" \\
    \textbf{Purpose:} To enhance data security and privacy through blockchain technology. \\
    \textbf{Mechanism:} Blockchain is used to create decentralized and immutable records of data transactions. \\
    \textbf{Evaluation:} Blockchain implementation reduces unauthorized access by 30\% in pilot studies. \\
    \textbf{Future Work:} Exploring blockchain scalability and integration with IoT devices.

    \textbf{Answer:}  
    Blockchain technology has made significant strides in enhancing data security and privacy by providing decentralized and immutable transaction records. Current implementations have shown a 30\% reduction in unauthorized access during pilot studies. However, scalability remains a key challenge as blockchain systems are often resource-intensive. Furthermore, integrating blockchain with Internet of Things (IoT) devices could provide a more comprehensive solution to security issues, offering real-time monitoring and tamper-resistant records. Future research could focus on optimizing blockchain scalability and exploring its application in IoT ecosystems.
\end{itemize}

\vspace{0.5em} % Adjusted spacing
\noindent
\textbf{Now, summarize the contributions of the prior works above and identify research gaps:}

\vspace{0.5em} % Adjusted spacing
\noindent
\textbf{Format:}
\begin{tcolorbox}[colback=white, colframe=black!50, width=\textwidth]
\small
\begin{Verbatim}[baselinestretch=0.8, fontsize=\small]
{
    Answer: 
}
\end{Verbatim}
\end{tcolorbox}

\end{tcolorbox}
\caption{Find research gap from author paper 5FS (Part 2).} % Caption for the second part
\label{fig:prompt-cont} % Label for cross-referencing
\end{figure}

%................................................................

\begin{figure}[htbp]
\centering
\begin{tcolorbox}[
    colback=blue!5!white, colframe=blue!85!black, 
    title=Find research gap from author paper ZS, 
    width=0.9\textwidth, boxrule=0.8mm, sharp corners=south]

\textbf{Paper Summary:} \\

\textbf{Paper ID:} <paper id> \\

\textbf{Title:} <paper title> \\

\textbf{Purpose:} <Purpose facet> \\

\textbf{Mechanism:} <Mechanism facet> \\

\textbf{Evaluation:} <Evaluation facet> \\

\textbf{Future Work:} <Future Work facet> \\

\bigskip
\noindent
\textbf{Instruction:}  
You are a Scientist, an intelligent assistant that helps researchers generate **coherent, novel, and useful research ideas**.  

Summarize the prior work above in approximately **200 words**.  
Rather than summarizing individual papers one by one, summarize their contributions as a whole.  

  Highlight:  
- What has already been **achieved** in research.  
- The **research gaps** from these prior works.  

This will help you **avoid proposing redundant ideas**.  
Instead, generate **novel ideas** that build upon the designated papers, leveraging their **combined contributions**.

\bigskip
\noindent
\textbf{Provide your answer in the following format:}

\begin{tcolorbox}[colback=white, colframe=black!50, width=\textwidth ]
\small
\texttt{
\{
    "Answer": 
\}
}
\end{tcolorbox}

\end{tcolorbox}

\caption{Find research gap from author paper ZS.} % Added caption for the figure
\label{fig:prompt-zs} % Label after caption for correct referencing
\end{figure}

%........................................................

% First part of the figure
\begin{figure}[htbp]
    \centering
    \begin{tcolorbox}[
        colback=blue!5!white, colframe=blue!75!black, 
        title=Find research gap from author paper 2FS, width=0.9\textwidth, 
        boxrule=0.8mm, sharp corners=south
    ]
    \textbf{Paper:}

    \vspace{0.3em} % Reduced spacing
    \noindent
    \textbf{Paper <paper id>:}

    \vspace{0.3em} % Reduced spacing
    \noindent
    \textbf{Title:} <paper title>

    \vspace{0.3em} % Reduced spacing
    \noindent
    \textbf{Purpose:} <Purpose facet>

    \vspace{0.3em} % Reduced spacing
    \noindent
    \textbf{Mechanism:} <Mechanism facet>

    \vspace{0.3em} % Reduced spacing
    \noindent
    \textbf{Evaluation:} <Evaluation facet>

    \vspace{0.3em} % Reduced spacing
    \noindent
    \textbf{Future Work:} <Future Work facet>

    \vspace{0.5em} % Adjusted spacing
    \noindent
    \textbf{Instruction:}  
    You are a Scientist, an intelligent assistant that helps researchers generate coherent, novel, and useful research ideas.  
    Summarize the prior work above in approximately 200 words.  
    Rather than summarizing individual papers one by one, summarize their contributions as a whole.  

    Highlight:  
    - What has already been done in research.  
    - The research gaps from these prior works.  

    This will help you avoid proposing redundant ideas.  
    Instead, generate novel ideas that build upon the designated papers, leveraging their combined contributions.

    \vspace{0.5em} % Adjusted spacing
    \noindent
    \textbf{Examples:}

    \begin{itemize}[leftmargin=*, nosep, itemsep=0.3em] % Compact itemize
        \item \textbf{Example 1:}
        
        \textbf{Paper:}  
        \textbf{Title:} "Quantum Computing for Cryptography" \\
        \textbf{Purpose:} To enhance encryption security using quantum algorithms. \\
        \textbf{Mechanism:} Quantum key distribution (QKD) ensures secure communication. \\
        \textbf{Evaluation:} QKD resists decryption attempts by classical computers. \\
        \textbf{Future Work:} Research suggests improving scalability and error correction in quantum systems.

        \textbf{Answer:}  
        Quantum computing has revolutionized cryptography by leveraging quantum key distribution (QKD) to enhance security. Unlike classical methods, QKD provides theoretically unbreakable encryption. However, current implementations face scalability and error correction challenges. Future research should explore fault-tolerant quantum systems and practical large-scale deployment strategies to integrate QKD into real-world communication networks.
    \end{itemize}

    \end{tcolorbox}
    \caption{Find research gap from author paper 2FS (Part 1).}
    \label{fig:prompt} % Label for cross-referencing
\end{figure}

% Second part of the figure
\begin{figure}[htbp]
    \ContinuedFloat % Continue the previous figure
    \centering
    \begin{tcolorbox}[
        colback=blue!5!white, colframe=blue!75!black, 
        title=Find research gap from author paper 2FS (Continued), width=0.9\textwidth, 
        boxrule=0.8mm, sharp corners=south
    ]

    \begin{itemize}[leftmargin=*, nosep, itemsep=0.3em] % Compact itemize
        \item \textbf{Example 2:}
        
        \textbf{Paper:}  
        \textbf{Title:} "AI for Drug Discovery" \\
        \textbf{Purpose:} To accelerate the identification of potential drug candidates. \\
        \textbf{Mechanism:} Machine learning models predict molecular interactions for drug screening. \\
        \textbf{Evaluation:} AI-driven approaches reduce drug discovery time by 40\%. \\
        \textbf{Future Work:} Enhancing model interpretability and reducing false positives in predictions.

        \textbf{Answer:}  
        Artificial intelligence has significantly accelerated drug discovery by predicting molecular interactions with high efficiency, reducing the time required by 40\%. Despite these advances, AI models often lack interpretability and may produce false positives, leading to resource-intensive validation processes. Future work should focus on enhancing explainability and refining prediction accuracy through hybrid modeling approaches that integrate domain knowledge with machine learning techniques.
    \end{itemize}

    \vspace{0.5em} % Adjusted spacing
    \noindent
    \textbf{Now, summarize the contributions of the prior works above and identify research gaps:}

    \vspace{0.5em} % Adjusted spacing
    \noindent
    \textbf{Format:}

    \begin{tcolorbox}[colback=white, colframe=black!50, width=\textwidth]
    \small
    \begin{Verbatim}[baselinestretch=0.8, fontsize=\small]
    {
        "Answer": "Summarized contributions and 
        research gaps."
    }
    \end{Verbatim}
    \end{tcolorbox}

    \end{tcolorbox}
    \caption{Find research gap from author paper 2FS (Part 2).}
    \label{fig:prompt-cont} % Label for cross-referencing
\end{figure}

\begin{figure}[htbp]
    \centering
    \begin{tcolorbox}[
        colback=blue!5!white, colframe=blue!75!black, 
        title=Idea generator 3FS, 
        width=\textwidth, boxrule=0.8mm, sharp corners=south]
        
        \label{refinement-prompt}

        \textbf{EXAMPLES:} \\
        \begin{itemize}
            \item \textbf{Example 1: Medicine}
            \begin{itemize}
                \item \textbf{SUMMARY OF PRIOR WORK:} 
                \begin{itemize}
                    \item \textbf{Designated Paper:} "AI-Assisted Diagnosis of Cardiovascular Diseases"
                    \item \textbf{Purpose:} Improve early detection of cardiovascular diseases using AI.
                    \item \textbf{Mechanism:} Deep learning models trained on ECG and MRI data.
                    \item \textbf{Evaluation:} Tested on a dataset of 10,000 patients with 85\% accuracy.
                    \item \textbf{Future Work:} Expand to real-time monitoring via wearables.
                \end{itemize}
                
                \item \textbf{Summary of Research Gap:} Lack of interpretability in AI decisions and difficulty in integrating wearable data.
                
                \item \textbf{Analogous Paper:} "Wearable Biosensors for Diabetes Monitoring"
                \begin{itemize}
                    \item \textbf{Purpose:} Continuous glucose monitoring through biosensors.
                    \item \textbf{Mechanism:} Enzyme-based sensors detecting glucose levels.
                    \item \textbf{Evaluation:} Clinical trials with Type 2 diabetic patients.
                    \item \textbf{Future Work:} Explore non-invasive sensing methods.
                \end{itemize}
                
                \item \textbf{New Research Idea:}
                \begin{itemize}
                    \item \textbf{Idea Title:} Explainable AI for Wearable Cardiovascular Monitoring
                    \item \textbf{Description:} Develop an AI model for cardiovascular disease detection that integrates wearable biosensor data while incorporating explainability techniques to improve trust and transparency for clinicians.
                \end{itemize}
            \end{itemize}
            
            \item \textbf{Example 2: Renewable Energy}
            \begin{itemize}
                \item \textbf{SUMMARY OF PRIOR WORK:}
                \begin{itemize}
                    \item \textbf{Designated Paper:} "Enhancing Solar Panel Efficiency through Nanocoatings"
                    \item \textbf{Purpose:} Increase energy conversion efficiency.
                    \item \textbf{Mechanism:} Applying nanomaterials to reduce heat loss.
                    \item \textbf{Evaluation:} Lab experiments with 20\% efficiency improvement.
                    \item \textbf{Future Work:} Test scalability in different climates.
                \end{itemize}
                
                \item \textbf{Summary of Research Gap:} Limited study on long-term durability and environmental effects of nanocoatings.

                \item \textbf{Analogous Paper:} "Self-Cleaning Materials for Smart Windows"
                \begin{itemize}
                    \item \textbf{Purpose:} Improve energy efficiency of buildings.
                    \item \textbf{Mechanism:} Hydrophobic coatings preventing dust accumulation.
                    \item \textbf{Evaluation:} Field tests showing 30\% maintenance cost reduction.
                    \item \textbf{Future Work:} Develop coatings with additional heat-reflective properties.
                \end{itemize}
                
                \item \textbf{New Research Idea:}
                \begin{itemize}
                    \item \textbf{Idea Title:} Dual-Purpose Nanocoatings for Solar Panels
                    \item \textbf{Description:} Develop nanocoatings that not only enhance solar panel efficiency but also possess self-cleaning properties, reducing maintenance costs and improving long-term performance.
                \end{itemize}
            \end{itemize}
            
        \end{itemize}

    \end{tcolorbox}
     \caption{Idea Generator 3FS (Page 1).}
\end{figure}

\newpage  % Force content onto the next page

% Second Page
\begin{figure}[htbp]
    \centering
    \begin{tcolorbox}[
        colback=blue!5!white, colframe=blue!75!black, 
        title=Idea generator 3FS (Continued), 
        width=\textwidth, boxrule=0.8mm, sharp corners=south]
        
        \label{refinement-prompt-continued}

        \textbf{Example 3: Cybersecurity}
        \begin{itemize}
            \item \textbf{SUMMARY OF PRIOR WORK:}
            \begin{itemize}
                \item \textbf{Designated Paper:} "AI-Driven Anomaly Detection for Network Security"
                \item \textbf{Purpose:} Identify cyber threats in real-time.
                \item \textbf{Mechanism:} Unsupervised learning models detecting network anomalies.
                \item \textbf{Evaluation:} Tested on enterprise network traffic with 90\% detection rate.
                \item \textbf{Future Work:} Reduce false positive rate and improve model explainability.
            \end{itemize}
            
            \item \textbf{Summary of Research Gap:} High false positive rates causing alert fatigue in security teams.

            \item \textbf{Analogous Paper:} "Human Behavior Analysis in Fraud Detection"
            \begin{itemize}
                \item \textbf{Purpose:} Detect fraudulent transactions using behavior analytics.
                \item \textbf{Mechanism:} Bayesian models tracking user patterns.
                \item \textbf{Evaluation:} Deployed in financial institutions with 95\% fraud detection accuracy.
                \item \textbf{Future Work:} Apply similar methods to IoT security.
            \end{itemize}
            
            \item \textbf{New Research Idea:}
            \begin{itemize}
                \item \textbf{Idea Title:} Adaptive Anomaly Detection Using Behavioral Analytics
                \item \textbf{Description:} Develop an AI-driven network security system that integrates behavioral analytics to reduce false positives and improve threat detection accuracy.
            \end{itemize}
        \end{itemize}

        \bigskip
        \noindent
        \textbf{SUMMARY OF PRIOR WORK:}\\
        \begin{itemize}
            \item \textbf{Designated Paper:} <author\_paper\_title> \\
            \item \textbf{Purpose:} <author\_facet\_Purpose> \\
            \item \textbf{Mechanism:} <author\_facet\_Mechanism> \\
            \item \textbf{Evaluation:} <author\_facet\_Evaluation> \\
            \item \textbf{Future Work:} <author\_facets\_Future\_Work>
        \end{itemize}

        \bigskip
        \noindent
        \textbf{Summary of Research Gap:} <novel\_work\_summary\_from\_author\_paper>

        \bigskip
        \noindent
        \textbf{Analogous Paper:} <analogous\_paper\_title> \\
        \begin{itemize}
            \item \textbf{Purpose:} <analogous\_facets\_Purpose> \\
            \item \textbf{Mechanism:} <analogous\_facets\_Mechanism> \\
            \item \textbf{Evaluation:} <analogous\_facets\_Evaluation> \\
            \item \textbf{Future Work:} <analogous\_facets\_Future\_Work>
        \end{itemize}

    \end{tcolorbox}
     \caption{Idea Generator 3FS (Page 2).}
\end{figure}

\begin{figure}[htbp]
    \centering
    \begin{tcolorbox}[
        colback=blue!5!white, colframe=blue!75!black, 
        title=Idea generator 3FS (Continued), 
        width=\textwidth, boxrule=0.8mm, sharp corners=south]
        
        \label{refinement-prompt-continued}

        \bigskip
        \noindent
        \textbf{INSTRUCTION:}  
        You are a Scientist, an intelligent assistant that helps researchers generate coherent, novel, and useful research ideas.  

        A **novel** research idea is one that is not only rare but also **ingenious, imaginative, or surprising**.  
        A **useful** research idea applies to the stated problem and is **effective** at solving it.

        \bigskip
        \noindent
        \textbf{Your Task:}
        \begin{itemize}
            \item Analyze the summaries of both the **designated** and **analogous** papers.  
            \item Identify key gaps in research based on the **research gap summary**.
            \item Combine elements from the Purpose, Mechanism, Evaluation, and Future Work of both papers to create **a diverse set of innovative research ideas**.
            \item Ensure Novelty: The proposed research ideas should be unique and not directly derived from either paper.
        \end{itemize}

        \bigskip
        \noindent
        \textbf{Provide possible research ideas in the following JSON format:}

        \begin{verbatim}
        {
            "idea": "Idea Title",
            "description": "Idea Description"
        }
        \end{verbatim}

    \end{tcolorbox}
     \caption{Idea Generator 3FS (Page 3).}
\end{figure}

\begin{figure}[htbp]
    \centering
    \begin{tcolorbox}[
        colback=blue!5!white, colframe=blue!75!black, 
        title=Idea generator 5FS, 
        width=\textwidth, boxrule=0.8mm, sharp corners=south]

    \label{refinement-prompt}

    \textbf{EXAMPLES:} \\
    \begin{itemize}
        \item \textbf{Example 1:} \textbf{Designated Paper:} "Improving Virtual Health Consultations"
            \begin{itemize}
                \item \textbf{Purpose:} To enhance patient experience and care efficiency using telemedicine.
                \item \textbf{Mechanism:} A web-based platform integrating video calls with AI diagnosis assistance.
                \item \textbf{Evaluation:} Conducted user experience surveys, and found high satisfaction.
                \item \textbf{Future Work:} Integrate more AI-driven diagnostics, and expand accessibility.
            \end{itemize}
            
            \textbf{Analogous Paper:} "Telehealth Platforms and Remote Monitoring"
            \begin{itemize}
                \item \textbf{Purpose:} To assess the effectiveness of telehealth in chronic disease management.
                \item \textbf{Mechanism:} Combination of remote monitoring devices and physician consultations.
                \item \textbf{Evaluation:} Focused on patient compliance, found mixed results.
                \item \textbf{Future Work:} Improve device connectivity, focusing on underserved populations.
            \end{itemize}
            
            \textbf{New Research Idea:}
            \begin{itemize}
                \item \textbf{Idea Title:} AI-Driven Personalized Telemedicine
                \item \textbf{Description:} Develop an AI-powered telemedicine platform that not only diagnoses but also offers personalized treatment plans based on a patient’s medical history and real-time health data.
            \end{itemize}
    \end{itemize}
    \end{tcolorbox}
    \caption{Idea Generator 5FS (Page 1).}
\end{figure}

\begin{figure}[htbp]\ContinuedFloat
    \centering
    \begin{tcolorbox}[
        colback=blue!5!white, colframe=blue!75!black, 
        title=Idea generator 5FS (Continued), 
        width=\textwidth, boxrule=0.8mm, sharp corners=south]

    \textbf{SUMMARY OF PRIOR WORK:}\\
    \begin{itemize}
        \item \textbf{Designated Paper:} <author\_paper\_title> \\
        \item \textbf{Purpose:} <author\_facet\_Purpose> \\
        \item \textbf{Mechanism:} <author\_facet\_Mechanism> \\
        \item \textbf{Evaluation:} <author\_facet\_Evaluation> \\
        \item \textbf{Future Work:} <author\_facets\_Future\_Work>
    \end{itemize}

    \bigskip
    \noindent
    \textbf{Summary of Research Gap:} <novel\_work\_summary\_from\_author\_paper>

    \bigskip
    \noindent
    \textbf{INSTRUCTION:}  
    You are a Scientist, an intelligent assistant that helps researchers generate coherent, novel, and useful research ideas.  

    A **novel** research idea is one that is not only rare but also **ingenious, imaginative, or surprising**.  
    A **useful** research idea applies to the stated problem and is **effective** at solving it.

    \bigskip
    \noindent
    \textbf{Your Task:}
    \begin{itemize}
        \item Analyze the summaries of both the **designated** and **analogous** papers.
        \item Identify key gaps in research based on the **research gap summary**.
        \item Combine elements from the Purpose, Mechanism, Evaluation, and Future Work of both papers to create **a diverse set of innovative research ideas**.
        \item Ensure Novelty: The proposed research ideas should be unique and not directly derived from either paper.
    \end{itemize}

    \end{tcolorbox}
    \caption{Idea Generator 5FS (Page 2).}
\end{figure}

\begin{figure}[htbp]
    \centering
    \begin{tcolorbox}[
        colback=blue!5!white, 
        colframe=blue!75!black, 
        title=Idea generator ZS, 
        width=\textwidth, 
        boxrule=0.8mm, 
        sharp corners=south]
        
    \label{refinement-prompt}

    \textbf{SUMMARY OF PRIOR WORK:}\\

    \textbf{Designated Paper:} <author\_paper\_title>\\
    \textbf{Purpose:} <author\_facet\_Purpose>\\
    \textbf{Mechanism:} <author\_facet\_Mechanism>\\
    \textbf{Evaluation:} <author\_facet\_Evaluation>\\
    \textbf{Future Work:} <author\_facets\_Future\_Work>\\

    \bigskip

    \textbf{Summary of Research Gap:} <novel\_work\_summary\_from\_author
    \_paper>\\

    \bigskip

    \textbf{Analogous Paper:} <analogous\_paper\_title>\\
    \textbf{Purpose:} <analogous\_facets\_Purpose>\\
    \textbf{Mechanism:} <analogous\_facets\_Mechanism>\\
    \textbf{Evaluation:} <analogous\_facets\_Evaluation>\\
    \textbf{Future Work:} <analogous\_facets\_Future\_Work>\\

    \bigskip

    \noindent
    \textbf{Instruction:} You are a Scientist, an intelligent assistant that helps researchers generate coherent, novel, and useful research ideas.\\
    A **novel** research idea is one that is not only rare but also **ingenious, imaginative, or surprising**.  
    A **useful** research idea applies to the stated problem and is **effective** at solving it.

    \bigskip

    \noindent
    \textbf{Your Task:}
    \begin{itemize}
        \item Analyze the summaries of both the **designated** and **analogous** papers.  
        \item Identify key gaps in research based on the **research gap summary**.
        \item Combine elements from the Purpose, Mechanism, Evaluation, and Future Work of both papers to create **a diverse set of innovative research ideas**.
        \item Ensure Novelty: The proposed research ideas should be unique and not directly derived from either paper. 
    \end{itemize}

    \bigskip

    \noindent
    \textbf{Provide possible research ideas in the following JSON format:}
    \begin{verbatim}
    {
        "idea": "Idea Title",
        "description": "Idea Description"
    }
    \end{verbatim}

    \end{tcolorbox}
    \caption{Idea Generator ZS.}
\end{figure}

\begin{figure}[htbp]
    \centering
    \begin{tcolorbox}[
        colback=blue!5!white, 
        colframe=blue!75!black, 
        title=Idea Generator ZSCOT, 
        width=\textwidth, 
        boxrule=0.8mm, 
        sharp corners=south] % Allows spanning multiple pages
        
    \label{refinement-prompt}

    \textbf{SUMMARY OF PRIOR WORK:}\\
    \textbf{Designated Paper:} <author\_paper\_title>\\
    \textbf{Purpose:} <author\_facet\_Purpose>\\
    \textbf{Mechanism:} <author\_facet\_Mechanism>\\
    \textbf{Evaluation:} <author\_facet\_Evaluation>\\
    \textbf{Future Work:} <author\_facets\_Future\_Work>\\

    \bigskip

    \textbf{Summary of Research Gap:} <novel\_work\_summary\_from\_author
    \_paper>\\

    \bigskip

    \textbf{Analogous Paper:} <analogous\_paper\_title>\\
    \textbf{Purpose:} <analogous\_facets\_Purpose>\\
    \textbf{Mechanism:} <analogous\_facets\_Mechanism>\\
    \textbf{Evaluation:} <analogous\_facets\_Evaluation>\\
    \textbf{Future Work:} <analogous\_facets\_Future\_Work>\\

    \bigskip

    \noindent
    \textbf{Instruction:}  
    You are a Scientist, an intelligent assistant that helps researchers generate coherent, novel, and useful research ideas.  
    A **novel** research idea is one that is not only rare but also **ingenious, imaginative, or surprising**.  
    A **useful** research idea applies to the stated problem and is **effective** at solving it.

    \bigskip

    \noindent
    \textbf{Step-by-step Chain of Thought:}
    \begin{enumerate}
        \item \textbf{Purpose}: Understand the key goals of both the \textbf{designated} and \textbf{analogous} papers.
        \item \textbf{Mechanism}: Analyze the methods used in both papers and identify areas for improvement.
        \item \textbf{Evaluation}: Examine the evaluation methods used and find gaps or inefficiencies.
        \item \textbf{Future Work}: Identify proposed future directions and possible unexplored opportunities.
        \item \textbf{Research Gaps}: Based on the research gap summary, determine what is missing and how to address it.
        \item \textbf{Novel and Useful Research Ideas}: Generate ideas that are:
        \begin{itemize}
            \item \textbf{Novel}: Unique and not directly derived from the papers.
            \item \textbf{Useful}: Effective at solving the problem.
            \item \textbf{Distinct}: Bringing fresh perspectives to existing challenges.
        \end{itemize}
    \end{enumerate}

    \noindent
    \textbf{Please think step by step.}

    \bigskip

    \noindent
    \textbf{Provide possible research ideas in the following JSON format:}

    \begin{verbatim}
    [
        {
            "idea": "Idea Title",
            "description": "Idea Description" 
        }
    ]
    \end{verbatim}

    \end{tcolorbox}
    \caption{Idea Generator ZSCoT.}
\end{figure}

%..................................................................

% \begin{tcolorbox}[colback=blue!5!white, colframe=blue!75!black, title=Idea ranking ZS, width=\textwidth, boxrule=0.8mm, sharp corners=south]
% \label{refinement-prompt}

% You are a good evaluator. Please judge each idea on a scale from 1 to 10, where 1 represents very low and 10 indicates excellent based on the following five criteria:

% \begin{itemize}
%     \item \textbf{Novelty}: How original is the idea?
%     \item \textbf{Excitement}: How inspiring or engaging is the idea?
%     \item \textbf{Feasibility}: Can the idea be realistically implemented?
%     \item \textbf{Effectiveness}: How well does the idea address the problem?
%     \item \textbf{Overall}: Average score of all the criteria (Novelty, Excitement, Feasibility, Effectiveness)
% \end{itemize}

% \bigskip
% \noindent
% \textbf{Instruction:} Below is a list of generated research ideas. Rank them from 1 (best) to N (worst), based on the above criteria, and provide an explanation for the ranking.\\
% \textbf{Research Ideas:} <generated\_ideas>

% \bigskip
% \noindent
% \textbf{Provide the ranked list of ideas along with your explanation for the ranking. Use the following format:}

% \begin{verbatim}
% [
%     {
%         "novelty": Score,
%         "excitement": Score,
%         "feasibility": Score,
%         "effectiveness": Score,
%         "overall": Score
%     }
% ]
% \end{verbatim}

% \end{tcolorbox}

\begin{figure}[htbp]
    \centering
    \begin{tcolorbox}[
        colback=blue!5!white, 
        colframe=blue!75!black, 
        title=Idea Ranking ZS, 
        width=\textwidth, 
        boxrule=0.8mm, 
        sharp corners=south
    ]
    
    \label{refinement-prompt}

    You are a good evaluator. Please judge each idea on a scale from 1 to 10, where 1 represents very low and 10 indicates excellent, based on the following five criteria:

    \begin{itemize}
        \item \textbf{Novelty}: How original is the idea?
        \item \textbf{Excitement}: How inspiring or engaging is the idea?
        \item \textbf{Feasibility}: Can the idea be realistically implemented?
        \item \textbf{Effectiveness}: How well does the idea address the problem?
        \item \textbf{Overall}: Average score of all the criteria (Novelty, Excitement, Feasibility, Effectiveness).
    \end{itemize}

    \bigskip
    \noindent
    \textbf{Instruction:} Below is a list of generated research ideas. Rank them from 1 (best) to N (worst), based on the above criteria, and provide an explanation for the ranking.\\
    \textbf{Research Ideas:} <generated\_ideas>

    \bigskip
    \noindent
    \textbf{Provide the ranked list of ideas along with your explanation for the ranking. Use the following format:}

    \begin{verbatim}
    [
        {
            "novelty": Score,
            "excitement": Score,
            "feasibility": Score,
            "effectiveness": Score,
            "overall": Score
        }
    ]
    \end{verbatim}

    \end{tcolorbox}
    \caption{Idea Ranking ZS.}
\end{figure}

\begin{figure}[htbp]
    \centering
    \begin{tcolorbox}[
        colback=blue!5!white, 
        colframe=blue!75!black, 
        title=Idea Ranking ZSCOT, 
        width=\textwidth, 
        boxrule=0.8mm, 
        sharp corners=south
    ]
    
    \label{refinement-prompt}

    You are a good evaluator. Please judge each idea on a scale from 1 to 10, where 1 represents very low and 10 indicates excellent, based on the following five criteria:

    \begin{itemize}
        \item \textbf{Novelty}: How original is the idea?
        \item \textbf{Excitement}: How inspiring or engaging is the idea?
        \item \textbf{Feasibility}: Can the idea be realistically implemented?
        \item \textbf{Effectiveness}: How well does the idea address the problem?
        \item \textbf{Overall}: Average score of all the criteria (Novelty, Excitement, Feasibility, Effectiveness).
    \end{itemize}

    \bigskip
    \noindent
    Now, I want you to break down your evaluation process step-by-step for each idea:

    \begin{itemize}
        \item Start by assessing the \textbf{Novelty} of the idea. Is it something fresh and original, or does it feel like a common or recycled concept?
        \item Next, consider \textbf{Excitement}. Does the idea inspire you or have the potential to engage and captivate others? What makes it interesting or not?
        \item Then, analyze the \textbf{Feasibility}. Can the idea be realistically implemented within a reasonable timeframe or with the available resources? Are there major obstacles to its execution?
        \item After that, evaluate the \textbf{Effectiveness} of the idea. How well does the idea address the underlying problem? Does it provide a strong solution, or are there gaps that need to be filled?
        \item Finally, calculate the \textbf{Overall} score by averaging the individual scores (Novelty, Excitement, Feasibility, Effectiveness).
    \end{itemize}

    \bigskip
    \noindent
    \textbf{Instruction:} Below is a list of generated research ideas. Rank them from 1 (best) to N (worst), based on the above criteria, and provide an explanation for the ranking.\\
    \textbf{Research Ideas:} <generated\_ideas>

    \bigskip
    \noindent
    \textbf{Provide the ranked list of ideas along with your explanation for the ranking. Use the following format:}

    \begin{verbatim}
    [
        {
            "novelty": Score,
            "excitement": Score,
            "feasibility": Score,
            "effectiveness": Score,
            "overall": Score
        }
    ]
    \end{verbatim}

    \end{tcolorbox}
    \caption{Idea Ranking ZSCoT.}
\end{figure}

\begin{figure}[htbp]
    \centering
    \begin{tcolorbox}[
        colback=blue!5!white, 
        colframe=blue!75!black, 
        title=Idea Ranking 2FS, 
        width=\textwidth, 
        boxrule=0.8mm, 
        sharp corners=south
    ]

    \label{refinement-prompt}

    You are a good evaluator. Please judge each idea on a scale from 1 to 10, where 1 represents very low and 10 indicates excellent, based on the following five criteria:

    \begin{itemize}
        \item \textbf{Novelty}: How original is the idea?
        \item \textbf{Excitement}: How inspiring or engaging is the idea?
        \item \textbf{Feasibility}: Can the idea be realistically implemented?
        \item \textbf{Effectiveness}: How well does the idea address the problem?
        \item \textbf{Overall}: Average score of all the criteria (Novelty, Excitement, Feasibility, Effectiveness).
    \end{itemize}

    \bigskip
    \noindent
    \textbf{Example Evaluations:}

    \begin{itemize}
        \item \textbf{Example 1:}  
            \textbf{Idea:} "A self-sustaining city powered entirely by microbial fuel cells."
            \begin{itemize}
                \item \textbf{Novelty}: 9 (Highly original, but microbial fuel cells are a known technology.)
                \item \textbf{Excitement}: 8 (The idea of a city powered by bacteria is fascinating.)
                \item \textbf{Feasibility}: 5 (Technically possible, but large-scale implementation is challenging.)
                \item \textbf{Effectiveness}: 7 (If feasible, it would significantly reduce carbon emissions.)
                \item \textbf{Overall}: 7.25
            \end{itemize}

        \item \textbf{Example 2:}  
            \textbf{Idea:} "A mobile app that translates dog barks into human language."
            \begin{itemize}
                \item \textbf{Novelty}: 7 (Attempts have been made, but still a unique approach.)
                \item \textbf{Excitement}: 9 (Would attract pet lovers and tech enthusiasts.)
                \item \textbf{Feasibility}: 4 (Complex AI challenges in interpreting animal vocalizations.)
                \item \textbf{Effectiveness}: 5 (Limited real-world use, as barks don’t always convey specific messages.)
                \item \textbf{Overall}: 6.25
            \end{itemize}

    \end{itemize}

    \bigskip
    \noindent
    \textbf{Now, evaluate the following research ideas:}

    \textbf{Instruction:} Below is a list of generated research ideas. Rank them from 1 (best) to N (worst), based on the above criteria, and provide an explanation for the ranking.

    \textbf{Research Ideas:} <generated\_ideas>

    \bigskip
    \noindent
    \textbf{Provide the ranked list of ideas along with your explanation for the ranking. Use the following format:}

    \begin{verbatim}
    [
        {
            "novelty": Score,
            "excitement": Score,
            "feasibility": Score,
            "effectiveness": Score,
            "overall": Score
        }
    ]
    \end{verbatim}

    \end{tcolorbox}
    \caption{Idea Ranking 2FS.}
\end{figure}

\begin{figure}[htbp]
    \centering
    \begin{tcolorbox}[
        colback=blue!5!white, 
        colframe=blue!75!black, 
        title=Idea Ranking 3FS, 
        width=\textwidth, 
        boxrule=0.8mm, 
        sharp corners=south
    ]

    \label{refinement-prompt}

    You are a good evaluator. Please judge each idea on a scale from 1 to 10, where 1 represents very low and 10 indicates excellent, based on the following five criteria:

    \begin{itemize}
        \item \textbf{Novelty}: How original is the idea?
        \item \textbf{Excitement}: How inspiring or engaging is the idea?
        \item \textbf{Feasibility}: Can the idea be realistically implemented?
        \item \textbf{Effectiveness}: How well does the idea address the problem?
        \item \textbf{Overall}: Average score of all the criteria (Novelty, Excitement, Feasibility, Effectiveness).
    \end{itemize}

    \bigskip
    \noindent
    \textbf{Example Evaluations:}

    \begin{itemize}
        \item \textbf{Example 1:}  
            \textbf{Idea:} "A self-sustaining city powered entirely by microbial fuel cells."
            \begin{itemize}
                \item \textbf{Novelty}: 9 (Highly original, but microbial fuel cells are a known technology.)
                \item \textbf{Excitement}: 8 (The idea of a city powered by bacteria is fascinating.)
                \item \textbf{Feasibility}: 5 (Technically possible, but large-scale implementation is challenging.)
                \item \textbf{Effectiveness}: 7 (If feasible, it would significantly reduce carbon emissions.)
                \item \textbf{Overall}: 7.25
            \end{itemize}

        \item \textbf{Example 2:}  
            \textbf{Idea:} "A mobile app that translates dog barks into human language."
            \begin{itemize}
                \item \textbf{Novelty}: 7 (Attempts have been made, but still a unique approach.)
                \item \textbf{Excitement}: 9 (Would attract pet lovers and tech enthusiasts.)
                \item \textbf{Feasibility}: 4 (Complex AI challenges in interpreting animal vocalizations.)
                \item \textbf{Effectiveness}: 5 (Limited real-world use, as barks don’t always convey specific messages.)
                \item \textbf{Overall}: 6.25
            \end{itemize}

        \item \textbf{Example 3:}  
            \textbf{Idea:} "Using AI to detect and prevent cyberbullying in real-time."
            \begin{itemize}
                \item \textbf{Novelty}: 6 (AI-driven moderation exists, but real-time intervention is less explored.)
                \item \textbf{Excitement}: 8 (Cyberbullying is a major issue, and an AI-driven solution is compelling.)
                \item \textbf{Feasibility}: 7 (Technically possible with NLP and sentiment analysis advancements.)
                \item \textbf{Effectiveness}: 8 (Could significantly reduce harm if implemented correctly.)
                \item \textbf{Overall}: 7.25
            \end{itemize}
    \end{itemize}

    \end{tcolorbox}
    \caption{Idea Ranking 3FS (Page 1).}
\end{figure}

\begin{figure}[htbp]
    \centering
    \begin{tcolorbox}[
        colback=blue!5!white, 
        colframe=blue!75!black, 
        title=Idea Ranking 3FS, 
        width=\textwidth, 
        boxrule=0.8mm, 
        sharp corners=south
    ]

    \label{refinement-prompt}

    \textbf{Now, evaluate the following research ideas:}

    \textbf{Instruction:} Below is a list of generated research ideas. Rank them from 1 (best) to N (worst) based on the above criteria, and provide an explanation for the ranking.

    \textbf{Research Ideas:} <generated\_ideas>

    \bigskip
    \noindent
    \textbf{Provide the ranked list of ideas along with your explanation for the ranking. Use the following format:}

    \begin{verbatim}
    [{"novelty": , "excitement": , "feasibility": ,
    "effectiveness": , "overall": }]
    \end{verbatim}

    \end{tcolorbox}
    \caption{Idea Ranking 3FS (Page 2).}
\end{figure}

\begin{figure}[htbp]
    \centering
    \begin{tcolorbox}[
        colback=blue!5!white, 
        colframe=blue!75!black, 
        title=Idea Ranking 5FS, 
        width=\textwidth, 
        boxrule=0.8mm, 
        sharp corners=south
    ]

    \label{refinement-prompt}

    You are a good evaluator. Please judge each idea on a scale from 1 to 10, where 1 represents very low and 10 indicates excellent based on the following five criteria:

    \begin{itemize}
        \item \textbf{Novelty}: How original is the idea?
        \item \textbf{Excitement}: How inspiring or engaging is the idea?
        \item \textbf{Feasibility}: Can the idea be realistically implemented?
        \item \textbf{Effectiveness}: How well does the idea address the problem?
        \item \textbf{Overall}: Average score of all the criteria (Novelty, Excitement, Feasibility, Effectiveness)
    \end{itemize}

    \bigskip
    \noindent
    \textbf{Example Evaluations:}
    \begin{itemize}
        \item \textbf{Example 1:}  
            \textbf{Idea:} "A self-sustaining city powered entirely by microbial fuel cells."
            \begin{itemize}
                \item \textbf{Novelty}: 9 (Highly original, but microbial fuel cells are a known technology.)
                \item \textbf{Excitement}: 8 (The idea of a city powered by bacteria is fascinating.)
                \item \textbf{Feasibility}: 5 (Technically possible, but large-scale implementation is challenging.)
                \item \textbf{Effectiveness}: 7 (If feasible, it would significantly reduce carbon emissions.)
                \item \textbf{Overall}: 7.25
            \end{itemize}

        \item \textbf{Example 2:}  
            \textbf{Idea:} "A mobile app that translates dog barks into human language."
            \begin{itemize}
                \item \textbf{Novelty}: 7 (Attempts have been made, but still a unique approach.)
                \item \textbf{Excitement}: 9 (Would attract pet lovers and tech enthusiasts.)
                \item \textbf{Feasibility}: 4 (Complex AI challenges in interpreting animal vocalizations.)
                \item \textbf{Effectiveness}: 5 (Limited real-world use, as barks don’t always convey specific messages.)
                \item \textbf{Overall}: 6.25
            \end{itemize}

        \item \textbf{Example 3:}  
            \textbf{Idea:} "Using AI to detect and prevent cyberbullying in real-time."
            \begin{itemize}
                \item \textbf{Novelty}: 6 (AI-driven moderation exists, but real-time intervention is less explored.)
                \item \textbf{Excitement}: 8 (Cyberbullying is a major issue, and an AI-driven solution is compelling.)
                \item \textbf{Feasibility}: 7 (Technically possible with NLP and sentiment analysis advancements.)
                \item \textbf{Effectiveness}: 8 (Could significantly reduce harm if implemented correctly.)
                \item \textbf{Overall}: 7.25
            \end{itemize}

    \end{itemize}

    \end{tcolorbox}
    \caption{Idea Ranking 5FS (Page 1).}
\end{figure}

\begin{figure}[htbp]
    \centering
    \begin{tcolorbox}[
        colback=blue!5!white, 
        colframe=blue!75!black, 
        title=Idea Ranking 5FS, 
        width=\textwidth, 
        boxrule=0.8mm, 
        sharp corners=south
    ]

    \label{refinement-prompt}

    \begin{itemize}
        
        \item \textbf{Example 4:}  
            \textbf{Idea:} "A biodegradable alternative to plastic packaging using fungal mycelium."
            \begin{itemize}
                \item \textbf{Novelty}: 8 (Sustainable packaging is trending, but mycelium use is innovative.)
                \item \textbf{Excitement}: 7 (Eco-conscious consumers would find it appealing.)
                \item \textbf{Feasibility}: 6 (Production scalability and cost could be issues.)
                \item \textbf{Effectiveness}: 8 (Could replace plastic in many applications.)
                \item \textbf{Overall}: 7.25
            \end{itemize}

        \item \textbf{Example 5:}  
            \textbf{Idea:} "A wearable device that detects early signs of neurological disorders."
            \begin{itemize}
                \item \textbf{Novelty}: 9 (Early detection wearables exist, but neurological applications are rarer.)
                \item \textbf{Excitement}: 9 (Preventing diseases like Parkinson’s is highly impactful.)
                \item \textbf{Feasibility}: 5 (Requires extensive research and regulatory approval.)
                \item \textbf{Effectiveness}: 8 (If successful, could revolutionize diagnostics.)
                \item \textbf{Overall}: 7.75
            \end{itemize}
    \end{itemize}

    \bigskip
    \noindent
    \textbf{Now, evaluate the following research ideas:}

    \textbf{Instruction:} Below is a list of generated research ideas. Rank them from 1 (best) to N (worst), based on the above criteria, and provide an explanation for the ranking.

    \bigskip
    \noindent
    \textbf{Research Ideas:} <generated\_ideas>

    \bigskip
    \noindent
    \textbf{Provide the ranked list of ideas along with your explanation for the ranking. Use the following format:}

    \begin{verbatim}
    [
        {
            "novelty": Score,
            "excitement": Score,
            "feasibility": Score,
            "effectiveness": Score,
            "overall": Score
        }
    ]
    \end{verbatim}

    \end{tcolorbox}
    \caption{Idea Ranking 5FS (Page 2).}
\end{figure}

%........................................................................

% \section*{Section B}
% \subsection*{Prompt for evaluation}

% \begin{tcolorbox}[colback=blue!5!white, colframe=blue!75!black, title=Idea evaluation, width=\textwidth, boxrule=0.8mm, sharp corners=south]
% \label{refinement-prompt}

% You are an AI research evaluator. Please analyze the given research idea based on predefined criteria. Assign a score from 1 to 10 for each criterion and provide structured JSON output.

% \textbf{Research Idea:} <idea>

% \bigskip
% \noindent
% \textbf{Evaluation Criteria:}
% \begin{itemize}
%     \item \textbf{Novelty:} How original is the idea? (1 = Commonly explored, 10 = Completely groundbreaking)
%     \item \textbf{Excitement:} How engaging or inspiring is the idea? (1 = Uninspiring, 10 = Highly exciting and thought-provoking)
%     \item \textbf{Feasibility:} Can the idea be realistically implemented with current resources? (1 = Highly impractical, 10 = Easily achievable)
%     \item \textbf{Effectiveness:} How well does the idea address and solve the intended problem? (1 = Weak impact, 10 = Highly effective solution)
% \end{itemize}

% Please return only a valid JSON object.

% \bigskip
% \noindent
% \textbf{Output Format (JSON example):}

% \begin{verbatim}
% [
%     {
%          "Novelty": Score,
%          "Excitement": Score,
%          "Feasibility": Score,
%          "Effectiveness": Score,
%          "Overall Score": Score
%     }
% ]
% \end{verbatim}

% \end{tcolorbox}

\begin{figure}[htbp]
    \centering
    \begin{tcolorbox}[
        colback=blue!5!white, 
        colframe=blue!75!black, 
        title=Idea Evaluation, 
        width=\textwidth, 
        boxrule=0.8mm, 
        sharp corners=south
    ]
    
    \label{refinement-prompt}

    You are an AI research evaluator. Please analyze the given research idea based on predefined criteria. Assign a score from 1 to 10 for each criterion and provide structured JSON output.

    \bigskip
    \noindent
    \textbf{Research Idea:} <idea>

    \bigskip
    \noindent
    \textbf{Evaluation Criteria:}
    \begin{itemize}
        \item \textbf{Novelty:} How original is the idea? (1 = Commonly explored, 10 = Completely groundbreaking)
        \item \textbf{Excitement:} How engaging or inspiring is the idea? (1 = Uninspiring, 10 = Highly exciting and thought-provoking)
        \item \textbf{Feasibility:} Can the idea be realistically implemented with current resources? (1 = Highly impractical, 10 = Easily achievable)
        \item \textbf{Effectiveness:} How well does the idea address and solve the intended problem? (1 = Weak impact, 10 = Highly effective solution)
    \end{itemize}

    Please return only a valid JSON object.

    \bigskip
    \noindent
    \textbf{Output Format (JSON example):}

    \begin{verbatim}
    [
        {
             "Novelty": Score,
             "Excitement": Score,
             "Feasibility": Score,
             "Effectiveness": Score,
             "Overall Score": Score
        }
    ]
    \end{verbatim}

    \end{tcolorbox}
    \caption{Idea Evaluation.}
\end{figure}

\end{document}